\documentclass[10pt,twocolumn,letterpaper]{article}

\usepackage[linesnumbered,ruled,lined,commentsnumbered]{algorithm2e}

\usepackage[dvipsnames]{xcolor}         % colors
\definecolor{linkColor}{rgb}{0.18,0.39,0.62} % Dark Skyblue.

\usepackage{microtype}
\usepackage{graphicx}
\usepackage{subfigure}
\usepackage{booktabs} % for professional tables
\usepackage{multirow}
\usepackage{pifont}
\usepackage{stmaryrd}
\usepackage{makecell}

\newcommand{\eeg}{{\it e.g., }}
\newcommand{\iie}{{\it i.e., }}

\definecolor{brickred}{rgb}{0.8, 0.25, 0.33}
\definecolor{brickgreen}{rgb}{0.25, 0.8, 0.33}
\newcommand{\cm}{\textcolor{brickgreen}{\ding{51}}}%
\newcommand{\xm}{\textcolor{brickred}{\ding{55}}}%
\usepackage{color}
\definecolor{brickred}{rgb}{0.8, 0.25, 0.33}
\definecolor{brickred2}{rgb}{0.25, 0.8, 0.33}
\usepackage{graphicx}

\newcommand{\algref}[1]{Algorithm~\ref{#1}}
\newcommand{\ttabref}[1]{Table~\ref{#1}}
\newcommand{\ffigref}[1]{Figure~\ref{#1}}
\newcommand{\ssecref}[1]{Section~\ref{#1}}
\newcommand{\eeqref}[1]{Equation~\ref{#1}}

\newcommand{\framework}{IDF}

\usepackage{iccv}              % To produce the CAMERA-READY version
\definecolor{iccvblue}{rgb}{0.21,0.49,0.74}
\usepackage[pagebackref,breaklinks,colorlinks,allcolors=iccvblue]{hyperref}

\title{IDF: Iterative Dynamic Filtering Networks for Generalizable Image Denoising}

\author{
Dongjin Kim$^{1}$\thanks{Equal contribution.} \quad
Jaekyun Ko$^{1}$\footnotemark[1] \quad
Muhammad Kashif Ali$^{2}$ \quad
Tae Hyun Kim$^{1}$\thanks{Corresponding author.} \\
$^{1}$Department of Computer Science, Hanyang University \\
$^{2}$School of Computing and Artificial Intelligence, Southwest Jiaotong University \\
{\tt\small \{dongjinkim, rhworbs1124, taehyunkim\}@hanyang.ac.kr \quad
kashifali@swjtu.edu.cn} \\
\tt\small \href{https://dongjinkim9.github.io/projects/idf}{https://dongjinkim9.github.io/projects/idf}
}

\begin{document}
\maketitle
\begin{abstract}
Image denoising is a fundamental challenge in computer vision, with applications in photography and medical imaging. 
While deep learning–based methods have shown remarkable success, their reliance on specific noise distributions limits generalization to unseen noise types and levels.
Existing approaches attempt to address this with extensive training data and high computational resources but they still suffer from overfitting.
To address these issues, we conduct image denoising by utilizing dynamically generated kernels via efficient operations.
This approach helps prevent overfitting and improves resilience to unseen noise. 
Specifically, our method leverages a Feature Extraction Module for robust noise-invariant features, Global Statistics and Local Correlation Modules to capture comprehensive noise characteristics and structural correlations. 
The Kernel Prediction Module then employs these cues to produce pixel-wise varying kernels adapted to local structures, which are then applied iteratively for denoising. This ensures both efficiency and superior restoration quality.
Despite being trained on single-level Gaussian noise, our compact model ($\sim$ 0.04 M) excels across diverse noise types and levels, demonstrating the promise of iterative dynamic filtering for practical image denoising.
\end{abstract}   
\vspace{-5mm}

\section{Introduction}
\label{sec:intro}

\begin{figure}
\begin{center}
\centerline{\includegraphics[width=1.00\columnwidth]{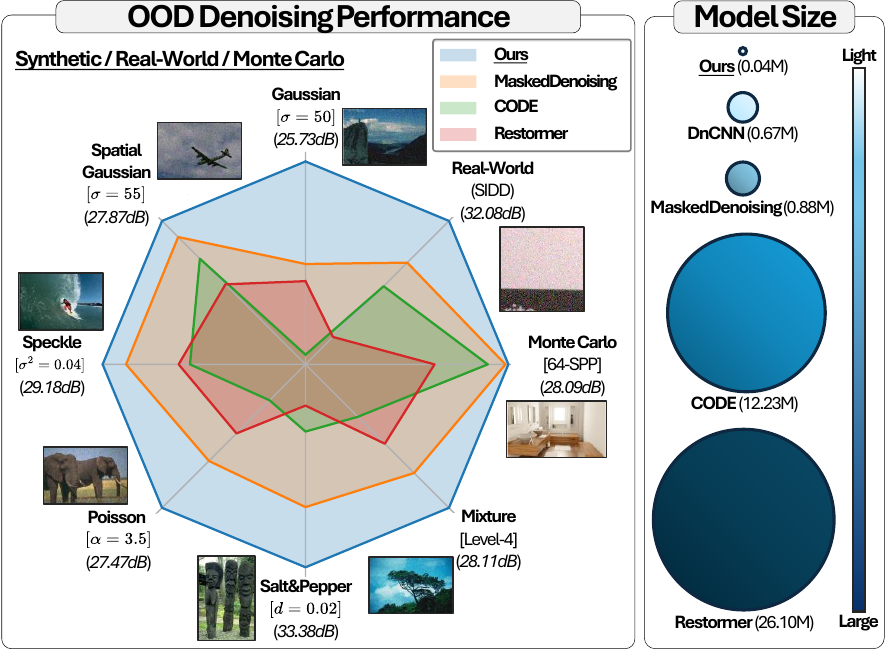}}
\caption{A comparison of denoising performance (PSNR) across various noise types and model sizes (number of parameters) between our method and comparable approaches~\cite{dncnn, masked_denoising, code, restormer}.}
\label{fig:teaser}
\vspace{-14mm}
\end{center}
\end{figure}

Image denoising remains a fundamental, yet challenging, problem in computer vision, with applications spanning photography and medical imaging. 
Recently, deep learning–based approaches~\cite{dncnn, ircnn, ffdnet, drunet, swinir, restormer, nafnet, hinet, uformer,mirnet,code} have demonstrated remarkable performance by learning an end-to-end mapping from noisy inputs to clean outputs. 
However, many of these methods are specifically tailored to the noise distributions present in their training datasets, limiting their generalization capabilities when faced with unseen noise types or levels~\cite{clip_denoising, masked_denoising,bias_free_network,drunet,dncnn}.

To address these challenges, several strategies have been proposed to improve robustness to out-of-distribution (OOD) noise. 
Self-supervised denoising networks~\cite{noise2noise, noise2self, apbsn, lgbpn, neighbor2neighbor, noise2void, high_quailty_denoising, random_subsample, mmbsn}, for instance, adapt neural networks to the unknown noise distribution without accessing clean images. However, these methods are vulnerable to noise types that are not included in the training data and often require a time-consuming test-time adaptation~\cite{gaintuning, lan, noiseda}. 
In contrast, various methods~\cite{noiseflow, naflow, danet, cycleisp, c2n, srgbflow, mfn} seek to mitigate overfitting issues by creating synthetic training data sets via noise modeling techniques. 
However, the resilience to OOD noise is still limited because the noise models are restricted to the training distribution.
While, prior-based methods~\cite{Ugpnet, diffbir, dps, clip_denoising} leverage priors from pretrained models trained on large training datasets to extract distortion-invariant features for robust denoising.
Nonetheless, these approaches are also limited in scenarios where extensive pretraining is not feasible. 
Recently, based on masked training strategies, the MaskedDenoising~\cite{masked_denoising} has been proposed to regularize networks to learn intrinsic image structures rather than overfitting to specific noise patterns. However, because masked training relies heavily on random masking, it can perform poorly in regions with fine-grained textures. %discriminative masking.

To overcome these limitations, we explore a lightweight dynamic kernel prediction approach, which offers a way to adapt filtering operations to the local image context. 
Unlike static convolutional filters, dynamic kernels are generated on a per-pixel basis, enabling the network to better capture spatially varying structures and noise patterns. 
In addition, we investigate iterative refinement strategies, which have demonstrated effectiveness in progressively enhancing predictions while preserving computational efficiency, and have been successfully employed in various low-level vision tasks~\cite{iterative_steering_kernel_regression, idr}.

Motivated by these approaches, we propose a novel \textbf{I}terative \textbf{D}ynamic \textbf{F}iltering network (\framework{}) that integrates dynamic kernel prediction with an adaptive iterative scheme. 
Our method is built on several key components. 
First, Kernel Prediction Module (KPM) generates pixel-wise varying denoising kernels and regularizes them by enforcing their elements to sum to one.
This sum-to-one constraint ensures that each kernel functions as a weighted averaging operator, guiding the model toward content-adaptive averaging rather than memorizing specific noise patterns.
In addition, Feature Extraction Module (FEM) applies sample-wise normalization that stabilizes feature statistics against unseen noise levels. 
Moreover, Global Statistics Module (GSM) and Local Correlation Module (LCM) add complementary global and local cues that improve kernel prediction. Collectively, these strategies prevent the memorization of training noise and encourage the learning of noise-invariant representations.
Finally, an adaptive iterative denoising strategy is employed, wherein the number of iterations is dynamically adjusted based on a confidence measure derived from the predicted kernels, ensuring both efficiency and high-quality denoising.

As summarized in \ffigref{fig:teaser}, despite being trained solely on Gaussian noise at a single level (\eeg  $\sigma=15$), our method outperforms conventional denoising approaches while maintaining a compact model size ($\sim$0.04M parameters). Specifically, \framework{} demonstrates effective generalization across various synthetic noise types, including Gaussian, Poisson, speckle, salt-and-pepper, spatially correlated Gaussian, and Monte Carlo-rendered image noise, as well as real-world noise captured by diverse sensors, 
such as smartphones and DSLR cameras.

\section{Related Works}
\label{related_works}
\subsection{Image Denoising}
Image denoising is a long-standing problem in image processing. 
Traditional denoising methods typically rely on small filters or hand-crafted priors~\cite{non_local_mean, anisotropic_diffusion, bm3d, total_variation, iterative_steering_kernel_regression}. 
For example, Non-Local Means~\cite{non_local_mean} exploits self-similarity and replaces each pixel value with a weighted average of similar patches across the image.
Similarly, BM3D~\cite{bm3d} clusters similar patches into 3D arrays and applies collaborative filtering in the transform domain, thereby leveraging non-local redundancy and sparsity. 
However, these prior-based methods have inherent limitations: their models are manually designed and rely solely on the given noisy image, rather than benefiting from large-scale external data. 

In contrast, modern approaches have adopted data-driven deep learning methods~\cite{dncnn, ircnn, ffdnet, drunet, swinir, restormer, nafnet, hinet, uformer,mirnet,code} where convolutional neural networks (CNNs) trained on extensive datasets of noisy-clean image pairs have achieved competitive results in noise removal.
More recently, transformers and state-space models have been introduced in the denoising task~\cite{swinir, restormer, mambair, mambairv2}.
For instance, methods such as SwinIR~\cite{swinir} and Restormer~\cite{restormer} employ self-attention mechanisms to capture long-range dependencies, thus boosting performance on benchmark datasets. 
Additionally, MambaIR~\cite{mambair} and MambaIRv2~\cite{mambairv2} adapt the selective structured state-space model to caputre long-range dependency with linear complexity.

\subsection{OOD Generalization}
Recent researches have explored generalizable image denoising methods that exhibit enhanced robustness to unseen noise~\cite{clip_denoising, masked_denoising,bias_free_network,drunet,dncnn,dil,hat,random_is_all_you_need}. 
One of these approaches involves training noise-blind models capable of handling a range of noise levels or types. 
For instance, DnCNN~\cite{dncnn} can be trained in a noise-blind manner on a mixture of noise levels to handle unknown noise.
Another promising direction is the masked denoising scheme~\cite{masked_denoising}, where random input patches are masked during training. This forces the network to reconstruct missing content from its surrounding context, encouraging the learning of intrinsic image structures rather than overfitting to specific noise patterns.
Recently, CLIPDenoising~\cite{clip_denoising} leverages features of the vision language model to enhance robustness. By incorporating dense features from a CLIP image encoder~\cite{clip} into the denoising network, this approach provides distortion-invariant and content-aware guidance that enables the model to generalize beyond the noise characteristics observed during training. 

\begin{figure*}[ht]
\begin{center}
\vspace{-5mm}
\centerline{\includegraphics[width=1.0\textwidth]{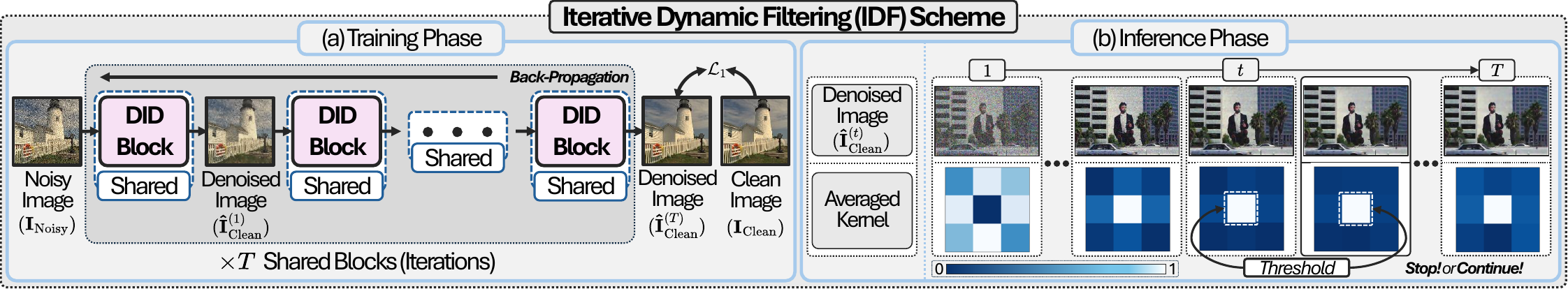}}
\caption{Overview of the proposed iterative dynamic filtering (IDF) scheme. (a) Training phase. (b) Inference phase.}
\label{fig:adaptive_iterative_scheme}
\vspace{-10mm}
\end{center}
\end{figure*}

\subsection{Dynamic Kernel Prediction}%Dynamic Filter Networks.}
Dynamic filter networks represent a class of methods in which filter weights are generated adaptively based on the input, rather than being fixed after training. Unlike conventional CNNs that apply the same learned kernels uniformly across all spatial locations, dynamic filter networks produce content-dependent kernels that enable position-specific and input-specific operations. This adaptability allows the network to tailor its processing to local patterns, a highly desirable property for vision tasks with spatially varying characteristics~\cite{kpcn,iterative_steering_kernel_regression,kpn,adfnet,df_video_interpolation, metasr}.

For instance, in the context of Monte Carlo-rendered image denoising, the Kernel-Predicting CNN~\cite{kpcn} introduces a novel approach that predicts spatially varying convolution kernels instead of directly outputting denoised pixel values. 
This method takes an advantage of additional scene information, such as depth, normals, and albedo, to enhance denoising performance. 
Similarly, for burst photography denoising, the Kernel Prediction Network~\cite{kpn} generates a three-dimensional denoising kernel for each pixel, effectively merging information from multiple noisy images. 
In sRGB image denoising, ADFNet~\cite{adfnet} employs a spatially enhanced kernel generator that captures a larger context when predicting each filter, thus addressing the limited spatial awareness of earlier dynamic filter approaches.

\subsection{Iterative Refinement}
Iterative refinement is a coarse-to-fine correction strategy 
that is widely used in numerous tasks~\cite{ddpm, stable_diffusion, iterative_steering_kernel_regression, sr3, diffir, rbpn, dsrn, dbpn, resshift, idr}.
A model begins with an initial estimate, measures the residual error, applies a learned update, and repeats this loop until convergence.
DBPN~\cite{dbpn} introduces an iterative up and down sampling module for super-resolution. 
The residuals generated by the back-projection layers capture global structures in early stages, 
while later iterations focus on high frequency details.
ResShift~\cite{resshift} employs a diffusion-driven paradigm for super-resolution. 
It forms a Markov chain that transfers between high- and low-resolution images by progressively shifting their residual, 
which improves transition efficiency and balances perceptual quality with reconstruction fidelity.
IDR~\cite{idr} addresses unsupervised image denoising by repeatedly removing noise 
while adding Gaussian perturbations at intermediate steps. 
This procedure enhances robustness on real-world noise benchmarks.
Leveraging these benefits, our work brings iterative refinement into the denoising domain, proposing an efficient and robust framework that progressively removes noise.

\section{Proposed Method}
\label{sec:methodology}

\subsection{Overall Flow}
\label{sec:3.1}

In \ffigref{fig:adaptive_iterative_scheme}, we illustrate the overall flow of~\framework{}. 
In both the training and inference phases, we employ an iterative denoising scheme. 
The input noisy image $\mathbf{I}_\mathrm{Noisy}$ is progressively denoised using $T$ Dynamic Image Denoising (DID) blocks to estimate the clean image $\mathbf{\hat{I}}^{(T)}_\mathrm{Clean}$, where the weights of these DID blocks are shared across all iterations to reduce the number of trainable parameters and avoid overfitting.
During inference, \framework{} offers two denoising strategies: fixed iteration and adaptive iteration. In the fixed iteration strategy, the noisy image is denoised for a predetermined number of iterations (fixed $T$), while the adaptive iteration strategy dynamically determines the number of iterations.
As illustrated in \ffigref{fig:adaptive_iterative_scheme} (b), an averaged kernel (confidence map) derived from the predicted kernels is used to dynamically adjust the number of iterations based on the complexity of the input noise and the image content. 
This map incorporates information on the convergence of the iterative denoising process.
This adaptive approach can accelerate inference speed by early terminating the iterative process once sufficient denoising has been achieved, effectively reducing computational overhead while maintaining comparable denoising performance.

\begin{figure*}[ht]
\begin{center}
\vspace{-5mm}
\centerline{\includegraphics[width=2.0\columnwidth]{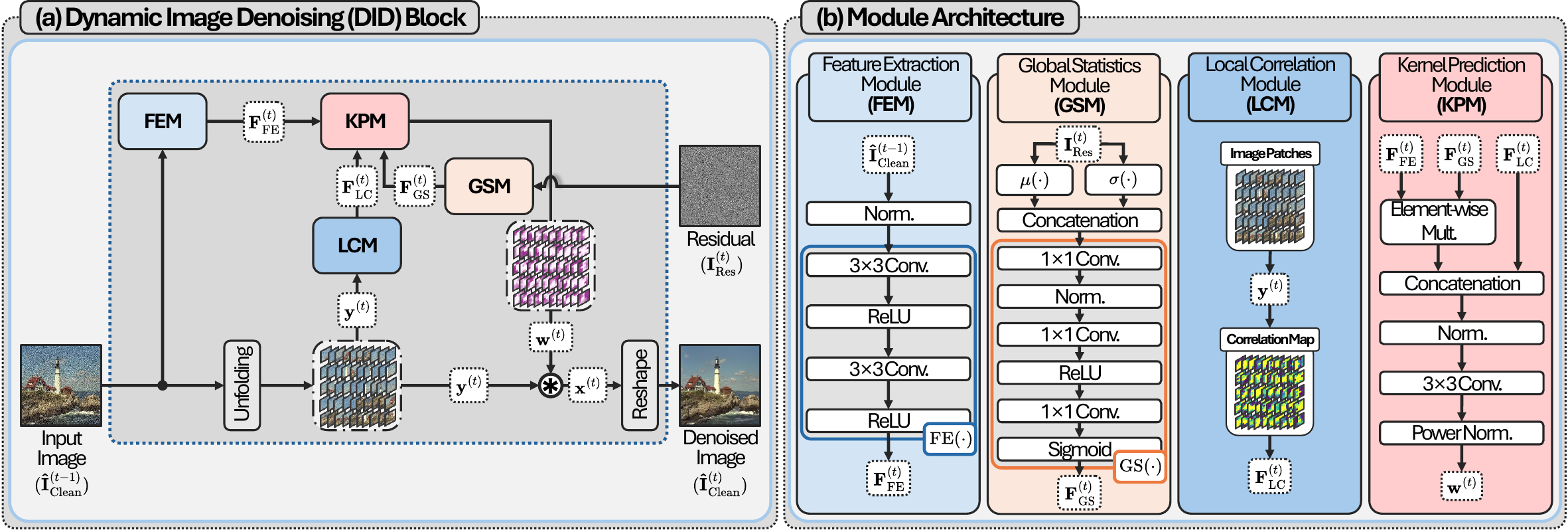}}
\caption{Architectural overview of the proposed method. (a) Dynamic image denoising (DID) block. (b) Module architectures.}
\label{fig:did_block}
\vspace{-10mm}
\end{center}
\end{figure*}

\subsection{Dynamic Image Denoising (DID)}
\label{sec:3.2}
As shown in \ffigref{fig:did_block}, the DID block estimates pixel-wise varying kernels that adaptively aggregate both global and local information to perform denoising. This process is repeated for $T$ iterations. 
Specifically, at the $t$-th iteration where $1 \leq t \leq T$, the previously denoised image $\hat{\mathbf{I}}_{\mathrm{Clean}}^{(t-1)}$ from the DID block is patchified by unfolding with a kernel of size $K\times K$ to generate overlapping patches, 
$\mathrm{\mathbf{y}}^{(t)} \in \mathbb{R}^{C \times K^2 \times (H \cdot W)}$, 
where $C$, $H$, and $W$ represent the number of channels, height, and width of the input image, respectively. These patches are used as the input of the DID block.
Notably, the input noisy image is initially used as  $\hat{\mathbf{I}}_{\mathrm{Clean}}^{(0)} = \mathrm{\mathbf{I}}_\mathrm{Noisy}$.
Moreover, during patch extraction at each iteration, we adjust the dilation rate, which controls the stride of elements within the neighborhood (\eeg  alternating between dilation rates of two and one) to encourage the learning of diverse features.

Then, using the input patches $\mathrm{\mathbf{y}}^{(t)}$,
the DID block predicts denoising kernels, $\mathrm{\mathbf{w}}^{(t)} \in \mathbb{R}^{1 \times K^2 \times (H \cdot W)}$, 
which are applied to the input patches and obtain the denoised output $\mathrm{\mathbf{x}}^{(t)}$.
Following the approach in~\cite{kpcn, kpn}, the same weights are applied uniformly across each RGB channel.
For a patch at a pixel location $i$, denoising is performed using the convolution operation as:
\begin{equation}
\mathrm{\mathbf{x}}^{(t)}_i = \mathrm{\mathbf{w}}_i^{(t)} \circledast \mathrm{\mathbf{y}}_i^{(t)},
\label{eq:patch_conv}
\end{equation}
where $\circledast$ represents the 2-D convolutional operation.
In the end, the denoised image is obtained by reshaping the overlapping patches $\mathrm{\mathbf{\hat{I}}}_\mathrm{Clean}^{(t)}$ as follows:
\begin{equation}
\label{eq:reshape}
\mathrm{\mathbf{\hat{I}}}_\mathrm{Clean}^{(t)} = \mathrm{Reshape}(\mathrm{\mathbf{x}}^{(t)}),
\end{equation}
where $\mathrm{Reshape}$ denotes the reshaping operation.

In particular, to dynamically generate the pixel-wise varying kernels, our DID block consists of four main components: the Feature Extraction Module (FEM), the Local Correlation Module (LCM), the Global Statistics Module (GSM), and the Kernel Prediction Module (KPM) as illustrated in \ffigref{fig:did_block} (b). We provide a detailed explanation of each module in the following.

\noindent\textbf{Feature Extraction Module (FEM).}
To predict pixel-wise varying kernels in DID, we first extract the features through the FEM. 
It is well established that improving the generalization performance of a denoising network relies on the extraction of robust features that are invariant to unseen noise~\cite{masked_denoising, clip_denoising}. 
These feature extractors require extensive data to train and can suffer from overfitting if trained on limited noise distributions (\eeg Gaussian noise with $\sigma=15$).
To address these issues that arise from training on a limited noise distribution and design a robust feature extractor, we design the FEM to be shallow, but invariant to the overall noise level, ensuring a consistent feature range across diverse real-world noise conditions. 
Inspired by previous works~\cite{imagenet, progressive_gan, edm2, rms_norm}, we first employ a straightforward yet effective normalization technique: root mean square (RMS) normalization, which normalizes the magnitude of the input of the FEM.
Specifically, we normalize the input by the RMS over its channel and spatial dimensions as follows:
\begin{equation}
    \mathrm{Norm}(\mathbf{a})= \frac{\mathbf{a}}{\sqrt{\frac{1}{N}\,\sum_{i=1}^{N} \bigl(\mathbf{a}_{i}\bigr)^2} + \epsilon},
\end{equation}
where $\mathbf{a}$ denotes the input of the normalization, $N$ indicates the total number of input elements (\eeg $CHW$), and a small constant $\epsilon$ is set to $10^{-4}$ for numerical stability.
Through this sample-wise normalization approach, the FEM becomes robust to the global noise-level changes.

Following this normalization step, a shallow Feature Extractor (FE) extracts features. The FE consists of two sequential $3 \times 3$ convolutional layers, each followed by a ReLU activation function, ensuring both simplicity and efficiency, which is formulated as follows:
\begin{equation}
\mathbf{F}_{\mathrm{FE}}^{(t)}=\mathrm{FE}\big(\mathrm{Norm}\big(\mathbf{\hat{I}}_\mathrm{Clean}^{(t-1)}\big)\big),
\end{equation}
where $\mathbf{F}^{(t)}_{\mathrm{FE}}$ represents the FEM feature at iteration $t$.

\noindent\textbf{Global Statistics Module (GSM).}
In denoising, it is well known that global information, such as ISO, can significantly help improve performance~\cite{ffdnet, unpaired_denoising, transfer_denoising}. Therefore, in our GSM, our aim is to estimate such noise level information. Estimating the accurate noise level of an input noisy image is a challenging problem. However, our framework employs an iterative approach, which gradually removes noise, and we can estimate the amount of noise removed at the previous iteration step by comparing the difference between the input and output images. In GSM, we use this information to improve kernel prediction accuracy in subsequent iterations.

Specifically, to estimate the amount of noise removed in the previous step, we first compute the residual noise $\mathrm{\mathbf{I}}^{(t)}_{\mathrm{Res}}$ as follows:
\begin{equation}
\mathrm{\mathbf{I}}^{(t)}_{\mathrm{Res}} = \mathrm{\mathbf{\hat{I}}}^{(t-1)}_\mathrm{Clean} - \mathrm{\mathbf{\hat{I}}}^{(t-2)}_\mathrm{Clean}.
\end{equation}
Note that for the first iteration ($t = 1$), the GSM is not applicable because there is no residual available.

Subsequently, we compute the mean $\mu$ and standard deviation $\sigma$ from $\mathrm{\mathbf{I}}^{(t)}_{\mathrm{Res}}$ at each channel, which serve as global measures of noise statistics. In particular, these statistical features are further processed by the GS module including several $1 \times 1$ convolutions and normalization as illustrated in 
\ffigref{fig:did_block} (b), and produce GSM features $\mathrm{\mathbf{F}}^{(t)}_\mathrm{GS}$ as follows:
\begin{equation}
\mathbf{F}_{\mathrm{GS}}^{(t)}=\mathrm{GS}\big(\big\llbracket\mu(\mathbf{I}_\mathrm{Res}^{(t)}),\sigma(\mathbf{I}_\mathrm{Res}^{(t)})\big\rrbracket\big).
\end{equation}

\noindent\textbf{Local Correlation Module (LCM).}
As adaptive kernels are influenced by the local structure of input images, it is essential to guide the kernel predictor to concentrate on the structure present within the patch.
To this end, we propose the LCM, where the local correlation map is computed by evaluating the similarity among pixels within patches of the input image. 
Specifically, following the approach in previous works~\cite{apbsn, lgbpn}, 
we define the LCM feature $\mathbf{F}_{\mathrm{LC}}^{(t)}$ by computing the Pearson correlation coefficients between neighboring pixels and the center pixel within each patch $\mathrm{\mathbf{y}}^{(t)}$ at iteration $t$.
By incorporating local correlation information, 
the following kernel prediction module can more effectively differentiate between regions of high self-similarity (\eeg homogeneous areas) 
and structured regions, such as edges. 
In homogeneous regions, high correlation values indicate that a nearly uniform kernel is suitable for averaging out noise. 
In contrast, in regions with low correlation, the network learns to assign kernels with more selective weights, thereby preserving important details.

\noindent\textbf{Kernel Prediction Module (KPM).}
Based on the features extracted by FEM, GSM and LCM, our KPM outputs pixel-wise varying denoising kernels to leverage neighboring information within each patch. 
In KPM, we first apply a channel-attention-like mechanism~\cite{channel_attention} to the input feature $\mathrm{\mathbf{F}}^{(t)}_\mathrm{FE}$ by multiplying $\mathrm{\mathbf{F}}^{(t)}_\mathrm{GS}$ by $\mathrm{\mathbf{F}}^{(t)}_\mathrm{FE}$.
Subsequently, the resulting feature map is concatenated with the LCM features $\mathrm{\mathbf{F}}^{(t)}_\mathrm{LC}$, then the concatenated features are normalized to account for the magnitude differences between the input sources, followed by a $3 \times 3$ convolution applied to generate the kernel weights $\mathrm{\mathbf{w}}^{(t)}$. 
This process at iteration $t$ is expressed as follows:
\begin{equation}
\mathrm{\mathbf{w}}^{(t)} = \mathrm{Conv}_{3\times3}\Big(\mathrm{Norm}\Big(\llbracket \mathrm{\mathbf{F}}_\mathrm{FE}^{(t)} \odot \mathrm{\mathbf{F}}_\mathrm{GS}^{(t)},\, \mathrm{\mathbf{F}}_\mathrm{LC}^{(t)}\rrbracket\Big)\Big),
\end{equation}
where $\odot$ indicates element-wise multiplication.
 
Moreover, to regularize kernel representations, we impose an additional constraint on the denoising kernel. 
Following previous works~\cite{kpcn, kpn}, we regularize the kernels by enforcing their elements to sum to one.
This sum-to-one constraint makes each kernel a weighted-averaging operator; it preserves the mean signal intensity and therefore avoids brightness or color shifts when the network encounters noise statistics not seen during training. The constraint also lowers the degrees of freedom in the prediction space, discouraging the network from memorizing training noise patterns.

Instead of the exponential normalization methods used in~\cite{kpn, kpcn}, we employ power normalization~\cite{power_norm1, power_norm2, power_norm3, power_norm4} due to its lower sensitivity to outliers compared to exponential-based functions such as softmax, as well as its computational simplicity. 
Given an input kernel $\mathbf{w}^{(t)}\in\mathbb{R}^{1\times K^{2}\times (H\cdot\\W)}$, 
we apply power normalization to each of the $K^{2}$ kernel maps independently at every spatial index $j\in\{1,\dots,H{\cdot}W\}$:
\begin{equation}
\mathrm{PowerNorm}\!\bigl(\mathbf{w}^{(t)}\bigr)_{i,j}
=\frac{\lvert \mathbf{w}^{(t)}_{i,j}\rvert^{\,p}}
       {\sum_{k=1}^{K^{2}}\lvert \mathbf{w}^{(t)}_{k,j}\rvert^{\,p}+\eta},
\label{eq:power_norm}
\end{equation}
where $i\in\{1,\dots,K^2\}$ is the index of the kernel map, $p$ controls the sharpness of the kernels, and $\eta$ is set to $10^{-4}$, ensuring numerical stability.

In particular, larger values of $p$ produce impulse-like kernels and prevent over-smoothing; in our experiments, we use $p = 3$. 
Using power normalization, the output kernel $\mathrm{\mathbf{w}}^{(t)}$ is normalized, and our kernel functions as a weighted averaging operation.

Finally, the normalized denoising kernels $\mathrm{\mathbf{w}}^{(t)}$ are convolved with the noisy input patches $\mathrm{\mathbf{y}}^{(t)}$ at each location of the pixel, and the denoised patches are further processed to generate $\mathrm{\mathbf{\hat{I}}}_\mathrm{Clean}^{(t)}$ through \eeqref{eq:patch_conv} and \eeqref{eq:reshape}, respectively.

\subsection{Dynamic Iteration Control (DIC)}
\label{sec:3.3}
\noindent\textbf{Confidence-based Adaptive Denoising.}
As depicted in~\ffigref{fig:adaptive_iterative_scheme}, \framework{} can remove noise iteratively through empirically determined steps $T$. 
However, for input images with low noise, excessive iterations can be inefficient and may result in over-smoothed outputs.
% across different noise types or levels can be inefficient and even lead to over-smoothed outputs for spatially sparse noise or inefficient processing for low-intensity noise. 
Thus, we propose an adaptive iterative denoising strategy called Dynamic Iteration Control (DIC), which dynamically determines the number of iterations $T$ based on the image content and noise characteristics during the inference phase.
Specifically, we utilize the predicted kernels for the DIC approach. 
We observe that in the early stages of denoising, when noise levels are high, the kernel values are widely dispersed within the kernel. However, as denoising is completed, the kernel values change less with each iteration, and the center of the denoising kernel gradually converges to 1.
From this observation, we define a confidence map and a criterion for early termination as:
\vspace{-1.5mm}
\begin{equation}
\begin{gathered}
\mathrm{\mathbf{C}}^{(t)} = \mathrm{\mathbf{w}}^{(t)}(c_x,c_y) - \mathrm{\mathbf{w}}^{(t-1)}(c_x,c_y),\\
T \leftarrow t, \quad \mathrm{if~}  \frac{1}{H\cdot W}\left|\sum^M_{i=1} \mathrm{\mathbf{C}}_i^{(t)} \right|  < \kappa,   
\end{gathered}
\label{eq:stop_threshold}
\vspace{-1.5mm}
\end{equation}
where $(c_x,c_y)$ denotes the center pixel location of the predicted kernels, and  $\mathrm{\mathbf{C}}^{(t)}$ represents the confidence map which measures the difference between the kernel centers over consecutive iterations. $M$ denotes the total number of spatial positions. 
The criterion is determined using the spatially averaged confidence map, with a threshold
$\kappa$ serving as a stopping condition. When this criterion is met, denoising is terminated. 

In our experiment, a criterion that measures the distance between the input $\mathbf{{y}}^{(t)}$ and the output $\mathbf{x}^{(t)}$ in iteration $t$, such as the mean absolute error, remains sensitive to the image content even in the later stages of denoising. 
In contrast, the criterion based on denoising kernel information demonstrates greater effectiveness, robustness, and ease of determination.

The overall inference algorithm of \framework{} is provided in the Supplementary Material~\algref{alg:DIC}.

\begin{table*}[t]
  \centering
  \caption{Quantitative results of denoising performance on CBSD68, McMaster, Kodak24 and Urban100 with regard to varied synthetic OOD noises in terms of PSNR$\uparrow$ and SSIM$\uparrow$. All methods are trained with Gaussian noise with a level of $\sigma=15$. The symbol $\dagger$ denotes that our model utilizes the proposed DIC during inference. The best and second-best results are highlighted in \textbf{bold} and \underline{underline}.}
  \vspace{-2.5mm}
  \resizebox{0.95\textwidth}{!}{
    \begin{tabular}{rc|cc|cccccccccccccc}
    \toprule
    \multicolumn{1}{r}{\multirow{2}{*}{Noise Types}} & \multirow{2}{*}{Datasets} & \multicolumn{2}{c|}{ClipDenoising} & \multicolumn{2}{c}{DnCNN} & \multicolumn{2}{c}{SwinIR} & \multicolumn{2}{c}{Restormer} & \multicolumn{2}{c}{CODE} & \multicolumn{2}{c}{MaskedDenoising} & \multicolumn{2}{c}{\textbf{\textbf{Ours}}$^\dagger$} & \multicolumn{2}{c}{\textbf{Ours}} \\
\cmidrule(lr){3-4} \cmidrule(lr){5-6} \cmidrule(lr){7-8} \cmidrule(lr){9-10} \cmidrule(lr){11-12} \cmidrule(lr){13-14} \cmidrule(lr){15-16} \cmidrule(lr){17-18}           &       & \multicolumn{2}{c|}{PSNR$\uparrow$/SSIM$\uparrow$}  & \multicolumn{2}{c}{PSNR$\uparrow$/SSIM$\uparrow$}  & \multicolumn{2}{c}{PSNR$\uparrow$/SSIM$\uparrow$}  & \multicolumn{2}{c}{PSNR$\uparrow$/SSIM$\uparrow$}  & \multicolumn{2}{c}{PSNR$\uparrow$/SSIM$\uparrow$}  & \multicolumn{2}{c}{PSNR$\uparrow$/SSIM$\uparrow$}  & \multicolumn{2}{c}{PSNR$\uparrow$/SSIM$\uparrow$}  & \multicolumn{2}{c}{PSNR$\uparrow$/SSIM$\uparrow$} \\
    \midrule
    \multicolumn{1}{r}{\multirow{4}{*}{\makecell{Gaussian \\ $\sigma=50$}}}  &  CBSD68  & \multicolumn{2}{c|}{25.84/0.7300} & \multicolumn{2}{c}{15.87/0.2269} & \multicolumn{2}{c}{16.07/0.2273} & \multicolumn{2}{c}{20.15/0.3859} & \multicolumn{2}{c}{16.72/0.2475} & \multicolumn{2}{c}{20.95/0.4411} & \multicolumn{2}{c}{\underline{25.41}/\underline{0.6932}} & \multicolumn{2}{c}{\textbf{25.73}/\textbf{0.7188}} \\
           &  MCMaster  & \multicolumn{2}{c|}{25.52/0.6879} & \multicolumn{2}{c}{16.34/0.1968} & \multicolumn{2}{c}{16.47/0.1942} & \multicolumn{2}{c}{20.33/0.3392} & \multicolumn{2}{c}{17.11/0.2128} & \multicolumn{2}{c}{21.33/0.3914} & \multicolumn{2}{c}{\underline{25.37}/\underline{0.6589}} & \multicolumn{2}{c}{\textbf{25.68}/\textbf{0.6871}} \\
           &  Kodak24  & \multicolumn{2}{c|}{26.69/0.7307} & \multicolumn{2}{c}{15.70/0.1867} & \multicolumn{2}{c}{15.92/0.1874} & \multicolumn{2}{c}{20.63/0.3635} & \multicolumn{2}{c}{16.58/0.2053} & \multicolumn{2}{c}{20.92/0.3857} & \multicolumn{2}{c}{\underline{26.00}/\underline{0.6763}} & \multicolumn{2}{c}{\textbf{26.60}/\textbf{0.7224}} \\
           &  Urban100  & \multicolumn{2}{c|}{25.13/0.7561} & \multicolumn{2}{c}{16.08/0.2968} & \multicolumn{2}{c}{16.26/0.2977} & \multicolumn{2}{c}{19.69/0.4282} & \multicolumn{2}{c}{16.92/0.3194} & \multicolumn{2}{c}{21.10/0.5020} & \multicolumn{2}{c}{\underline{24.78}/\underline{0.7319}} & \multicolumn{2}{c}{\textbf{25.06}/\textbf{0.7547}} \\
    \midrule
    \multicolumn{1}{r}{\multirow{4}{*}{\makecell{Spatial \\ Gaussian \\ $\sigma=55$}}}  &  CBSD68  & \multicolumn{2}{c|}{27.57/0.8004} & \multicolumn{2}{c}{25.88/0.6985} & \multicolumn{2}{c}{25.13/0.6482} & \multicolumn{2}{c}{23.34/0.5975} & \multicolumn{2}{c}{25.17/0.6471} & \multicolumn{2}{c}{26.72/0.7685} & \multicolumn{2}{c}{\underline{27.70}/\underline{0.7971}} & \multicolumn{2}{c}{\textbf{27.87}/\textbf{0.8049}} \\
           &  MCMaster  & \multicolumn{2}{c|}{28.38/0.7820} & \multicolumn{2}{c}{26.42/0.6756} & \multicolumn{2}{c}{25.61/0.6209} & \multicolumn{2}{c}{23.20/0.5439} & \multicolumn{2}{c}{25.52/0.6021} & \multicolumn{2}{c}{27.17/0.7324} & \multicolumn{2}{c}{\underline{28.01}/\underline{0.7642}} & \multicolumn{2}{c}{\textbf{28.46}/\textbf{0.7921}} \\
           &  Kodak24  & \multicolumn{2}{c|}{28.17/0.7881} & \multicolumn{2}{c}{25.89/0.6541} & \multicolumn{2}{c}{25.12/0.5997} & \multicolumn{2}{c}{22.62/0.5445} & \multicolumn{2}{c}{25.22/0.6043} & \multicolumn{2}{c}{27.22/0.7555} & \multicolumn{2}{c}{\underline{28.11}/\underline{0.7755}} & \multicolumn{2}{c}{\textbf{28.48}/\textbf{0.7933}} \\
           &  Urban100  & \multicolumn{2}{c|}{27.73/0.8213} & \multicolumn{2}{c}{26.25/0.7331} & \multicolumn{2}{c}{25.43/0.6885} & \multicolumn{2}{c}{24.06/0.6386} & \multicolumn{2}{c}{25.40/0.6918} & \multicolumn{2}{c}{26.18/0.7988} & \multicolumn{2}{c}{\underline{27.54}/\underline{0.8217}} & \multicolumn{2}{c}{\textbf{27.78}/\textbf{0.8333}} \\
    \midrule
    \multicolumn{1}{r}{\multirow{4}{*}{\makecell{Poisson \\ $\alpha=3.5$}}}  &  CBSD68  & \multicolumn{2}{c|}{27.63/0.8179} & \multicolumn{2}{c}{19.36/0.4304} & \multicolumn{2}{c}{19.45/0.4202} & \multicolumn{2}{c}{22.23/0.5598} & \multicolumn{2}{c}{19.90/0.4267} & \multicolumn{2}{c}{24.17/0.6352} & \multicolumn{2}{c}{\underline{27.17}/\underline{0.7852}} & \multicolumn{2}{c}{\textbf{27.47}/\textbf{0.8006}} \\
           &  MCMaster  & \multicolumn{2}{c|}{28.79/0.8209} & \multicolumn{2}{c}{20.31/0.5164} & \multicolumn{2}{c}{20.34/0.5132} & \multicolumn{2}{c}{21.89/0.5791} & \multicolumn{2}{c}{20.75/0.4656} & \multicolumn{2}{c}{25.09/0.5836} & \multicolumn{2}{c}{\underline{28.60}/\underline{0.8146}} & \multicolumn{2}{c}{\textbf{28.74}/\textbf{0.8219}} \\
           &  Kodak24  & \multicolumn{2}{c|}{28.51/0.8095} & \multicolumn{2}{c}{19.12/0.3617} & \multicolumn{2}{c}{19.23/0.3537} & \multicolumn{2}{c}{22.50/0.5144} & \multicolumn{2}{c}{19.69/0.3599} & \multicolumn{2}{c}{24.21/0.5737} & \multicolumn{2}{c}{\underline{28.08}/\underline{0.7778}} & \multicolumn{2}{c}{\textbf{28.41}/\textbf{0.7978}} \\
           &  Urban100  & \multicolumn{2}{c|}{27.08/0.8373} & \multicolumn{2}{c}{19.11/0.4979} & \multicolumn{2}{c}{19.19/0.4931} & \multicolumn{2}{c}{21.22/0.5682} & \multicolumn{2}{c}{19.69/0.4968} & \multicolumn{2}{c}{23.82/0.6667} & \multicolumn{2}{c}{\underline{26.82}/\underline{0.8317}} & \multicolumn{2}{c}{\textbf{27.01}/\textbf{0.8417}} \\
    \midrule
    \multicolumn{1}{r}{\multirow{4}{*}{\makecell{Salt \\ \& \\ Pepper \\ $d=0.02$}}}  &  CBSD68  & \multicolumn{2}{c|}{29.82/0.8440} & \multicolumn{2}{c}{24.02/0.7084} & \multicolumn{2}{c}{23.23/0.6735} & \multicolumn{2}{c}{23.60/0.6791} & \multicolumn{2}{c}{24.10/0.6858} & \multicolumn{2}{c}{29.74/0.8431} & \multicolumn{2}{c}{\textbf{33.74}/\textbf{0.9216}} & \multicolumn{2}{c}{\underline{33.38}/\underline{0.9129}} \\
           &  MCMaster  & \multicolumn{2}{c|}{29.77/0.8064} & \multicolumn{2}{c}{22.98/0.6615} & \multicolumn{2}{c}{22.68/0.6392} & \multicolumn{2}{c}{23.03/0.6396} & \multicolumn{2}{c}{23.38/0.6259} & \multicolumn{2}{c}{29.25/0.7726} & \multicolumn{2}{c}{\textbf{32.83}/\textbf{0.8911}} & \multicolumn{2}{c}{\underline{32.45}/\underline{0.8839}} \\
           &  Kodak24  & \multicolumn{2}{c|}{30.48/0.8353} & \multicolumn{2}{c}{24.26/0.6785} & \multicolumn{2}{c}{23.42/0.6383} & \multicolumn{2}{c}{23.82/0.6447} & \multicolumn{2}{c}{24.42/0.6577} & \multicolumn{2}{c}{30.44/0.8429} & \multicolumn{2}{c}{\textbf{34.36}/\textbf{0.9102}} & \multicolumn{2}{c}{\underline{34.10}/\underline{0.9039}} \\
           &  Urban100  & \multicolumn{2}{c|}{29.64/0.8634} & \multicolumn{2}{c}{23.55/0.7336} & \multicolumn{2}{c}{22.90/0.7075} & \multicolumn{2}{c}{23.42/0.7144} & \multicolumn{2}{c}{24.07/0.7246} & \multicolumn{2}{c}{28.42/0.8613} & \multicolumn{2}{c}{\textbf{32.39}/\textbf{0.9199}} & \multicolumn{2}{c}{\underline{32.15}/\underline{0.9150}} \\
    \midrule
    \multicolumn{1}{r}{\multirow{4}{*}{\makecell{Speckle \\ ${\sigma}^2=0.04$}}}  &  CBSD68  & \multicolumn{2}{c|}{29.50/0.8698} & \multicolumn{2}{c}{24.55/0.6964} & \multicolumn{2}{c}{24.08/0.6812} & \multicolumn{2}{c}{25.16/0.7202} & \multicolumn{2}{c}{24.56/0.6893} & \multicolumn{2}{c}{27.94/0.8144} & \multicolumn{2}{c}{\underline{29.08}/\underline{0.8521}} & \multicolumn{2}{c}{\textbf{29.18}/\textbf{0.8553}} \\
           &  MCMaster  & \multicolumn{2}{c|}{30.47/0.8452} & \multicolumn{2}{c}{25.08/0.7044} & \multicolumn{2}{c}{24.46/0.6910} & \multicolumn{2}{c}{25.28/0.7167} & \multicolumn{2}{c}{25.15/0.6511} & \multicolumn{2}{c}{28.66/0.7361} & \multicolumn{2}{c}{\underline{30.24}/\underline{0.8557}} & \multicolumn{2}{c}{\textbf{30.28}/\textbf{0.8561}} \\
           &  Kodak24  & \multicolumn{2}{c|}{30.44/0.8722} & \multicolumn{2}{c}{24.75/0.6505} & \multicolumn{2}{c}{24.17/0.6269} & \multicolumn{2}{c}{25.58/0.6851} & \multicolumn{2}{c}{24.77/0.6414} & \multicolumn{2}{c}{28.59/0.8031} & \multicolumn{2}{c}{\underline{29.95}/\underline{0.8451}} & \multicolumn{2}{c}{\textbf{30.08}/\textbf{0.8505}} \\
           &  Urban100  & \multicolumn{2}{c|}{28.70/0.8746} & \multicolumn{2}{c}{23.28/0.6872} & \multicolumn{2}{c}{22.88/0.6772} & \multicolumn{2}{c}{24.16/0.7105} & \multicolumn{2}{c}{23.44/0.6819} & \multicolumn{2}{c}{26.64/0.8043} & \multicolumn{2}{c}{\underline{28.42}/\underline{0.8716}} & \multicolumn{2}{c}{\textbf{28.45}/\textbf{0.8740}} \\
    \midrule
    \multicolumn{1}{r}{\multirow{4}{*}{\makecell{Mixture \\ Level 4}}}  &  CBSD68  & \multicolumn{2}{c|}{28.36/0.8349} & \multicolumn{2}{c}{21.38/0.4754} & \multicolumn{2}{c}{21.26/0.4533} & \multicolumn{2}{c}{23.49/0.5730} & \multicolumn{2}{c}{21.55/0.4556} & \multicolumn{2}{c}{25.60/0.7171} & \multicolumn{2}{c}{\underline{27.91}/\underline{0.8130}} & \multicolumn{2}{c}{\textbf{28.11}/\textbf{0.8224}} \\
           &  MCMaster  & \multicolumn{2}{c|}{29.13/0.8030} & \multicolumn{2}{c}{21.95/0.4779} & \multicolumn{2}{c}{21.87/0.4626} & \multicolumn{2}{c}{23.66/0.5613} & \multicolumn{2}{c}{22.10/0.4499} & \multicolumn{2}{c}{26.65/0.6842} & \multicolumn{2}{c}{\underline{29.00}/\underline{0.8079}} & \multicolumn{2}{c}{\textbf{29.09}/\textbf{0.8116}} \\
           &  Kodak24  & \multicolumn{2}{c|}{29.59/0.8450} & \multicolumn{2}{c}{22.00/0.4352} & \multicolumn{2}{c}{21.76/0.4099} & \multicolumn{2}{c}{24.88/0.5670} & \multicolumn{2}{c}{22.07/0.4175} & \multicolumn{2}{c}{26.32/0.6980} & \multicolumn{2}{c}{\underline{28.86}/\underline{0.8116}} & \multicolumn{2}{c}{\textbf{29.09}/\textbf{0.8239}} \\
           &  Urban100  & \multicolumn{2}{c|}{27.80/0.8452} & \multicolumn{2}{c}{21.26/0.5215} & \multicolumn{2}{c}{21.16/0.5057} & \multicolumn{2}{c}{23.20/0.5990} & \multicolumn{2}{c}{21.50/0.5154} & \multicolumn{2}{c}{25.17/0.7461} & \multicolumn{2}{c}{\underline{27.37}/\underline{0.8328}} & \multicolumn{2}{c}{\textbf{27.52}/\textbf{0.8405}} \\
    \bottomrule
    \end{tabular}%
    }
  \label{tab:synthetic_denoising}%
\vspace{-2.5mm}
\end{table*}%

\section{Experiments}
\subsection{Experimental Setup}
\label{sec:4.1}
\noindent\textbf{Implementation Details.}
Our model is optimized with AdamW~\cite{adamw}, minimizing the $\mathcal{L}_{1}$ distance between the final estimate ${\mathbf{\hat{I}}}_\mathrm{Clean}^{(T)}$ and the ground-truth image ${\mathbf{I}}_{\mathrm{Clean}}$. 
Training proceeds for 50k iterations at a fixed learning rate of $1\times 10^{-4}$. We sample random $128 \times 128$ patches, apply horizontal and vertical flips for data augmentation, and use a mini-batch size of 8. 
The maximum number of iterations is set to $T=10$, and the threshold in \eeqref{eq:stop_threshold} for the adaptive control (DIC) is fixed to $\kappa = 0.015$.

\noindent\textbf{Dataset.}
Following~\cite{clip_denoising}, we use the CBSD432~\cite{cbsd} dataset to train~\framework{}, synthesizing noisy images by adding \textit{i.i.d.} random Gaussian noise with $\sigma=15$.
We then evaluate denoising quality on three noise categories: synthetic noise, real-world sensor noise, and Monte Carlo rendering noise, using the evaluation setups in \cite{masked_denoising, clip_denoising}.
For synthetic noise evaluation, we employ four benchmarks: CBSD68 \cite{cbsd}, McMaster \cite{mcmaster}, Kodak24 \cite{kodak}, and Urban100 \cite{urban}. Each dataset is corrupted with six noise models: Gaussian, spatially correlated Gaussian, Poisson, speckle, salt-and-pepper, and mixture noise.
For real-world noise, we use SIDD \cite{sidd}, SIDD+ \cite{siddplus}, PolyU \cite{polyu}, and Nam \cite{nam}, which contain images captured by a range of smartphone and DSLR sensors.
% Finally, Monte Carlo rendering noise is assessed with the test set introduced in \cite{mc_dataset}.
Finally, for Monte Carlo rendering noise, we employ the test dataset proposed by~\cite{mc_dataset}.
Further information on the noise categories and their intensity levels is provided in the Supplementary Material \ssecref{appendix:noise_details}.

\subsection{Generalizable Denoising Performance}
\noindent\textbf{Compared Methods.}
We evaluate the generalization capability of our denoiser against DnCNN \cite{dncnn}, SwinIR \cite{swinir}, Restormer \cite{restormer}, CODE \cite{code}, and MaskedDenoising \cite{masked_denoising}.
To ensure a fair comparison, we use the official pretrained weights for the baseline models, which were trained on \textit{i.i.d.} Gaussian noise with $\sigma = 15$, consistent with our training configuration.
For evaluation, Peak Signal-to-Noise Ratio (PSNR) and Structural Similarity Index Measure (SSIM)~\cite{ssim} are used.
Consistent with \cite{random_is_all_you_need}, we exclude a direct comparison with the current state-of-the-art (SOTA) CLIPDenoising \cite{clip_denoising}, because its CLIP backbone is pretrained on a large-scale dataset that already includes a wide variety of noise types.
This introduces potential bias when compared to models that are exclusively trained using only fixed Gaussian noise. 
Therefore, we refrain from directly comparing with the CLIPDenoising model, but include its results for reference.

\begin{figure*}[ht]
\begin{center}
\centerline{\includegraphics[width=1.55\columnwidth]{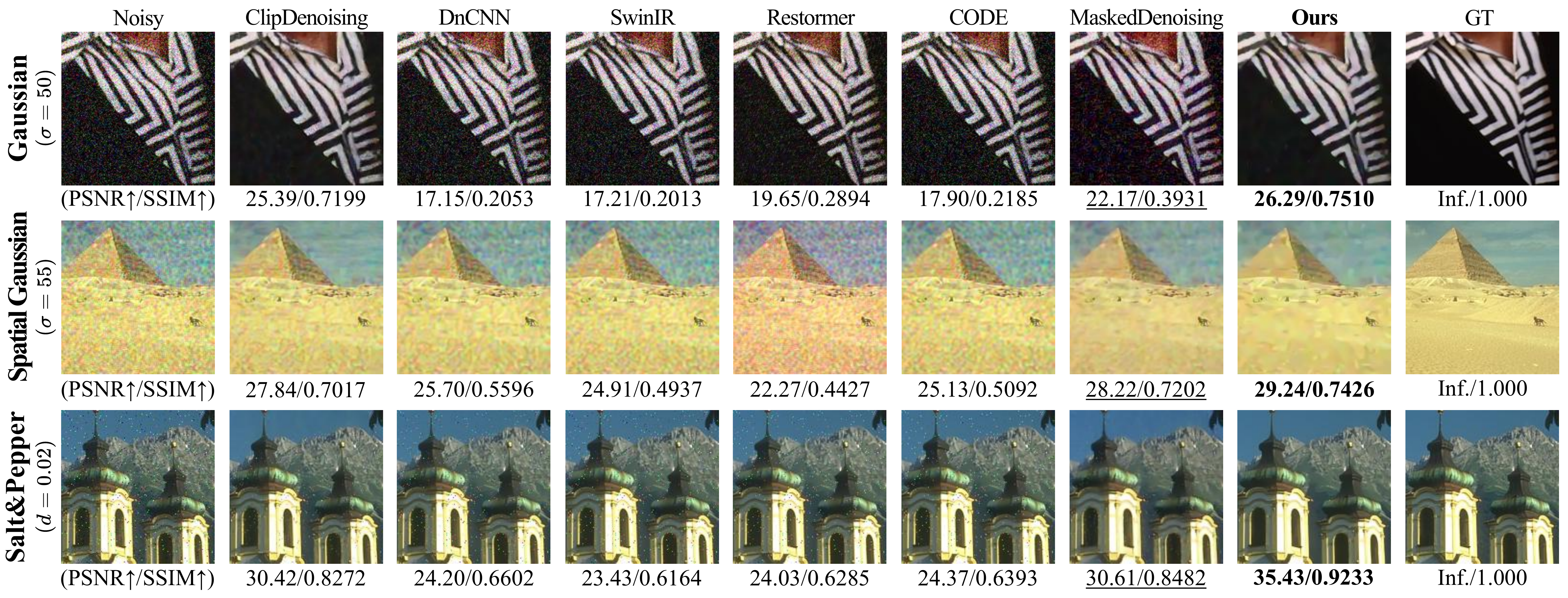}}
\caption{Qualitative results of denoising performance on synthetic OOD noise in terms of PSNR$\uparrow$/SSIM$\uparrow$. Our method consistently restores cleaner textures and sharper edges, especially in high-frequency regions, while suppressing residual artifacts seen in other methods.}
\label{fig:synthetic_denoising}
\vspace{-11mm}
\end{center}
\end{figure*}

\begin{figure}
\begin{center}
\centerline{\includegraphics[width=1\columnwidth]{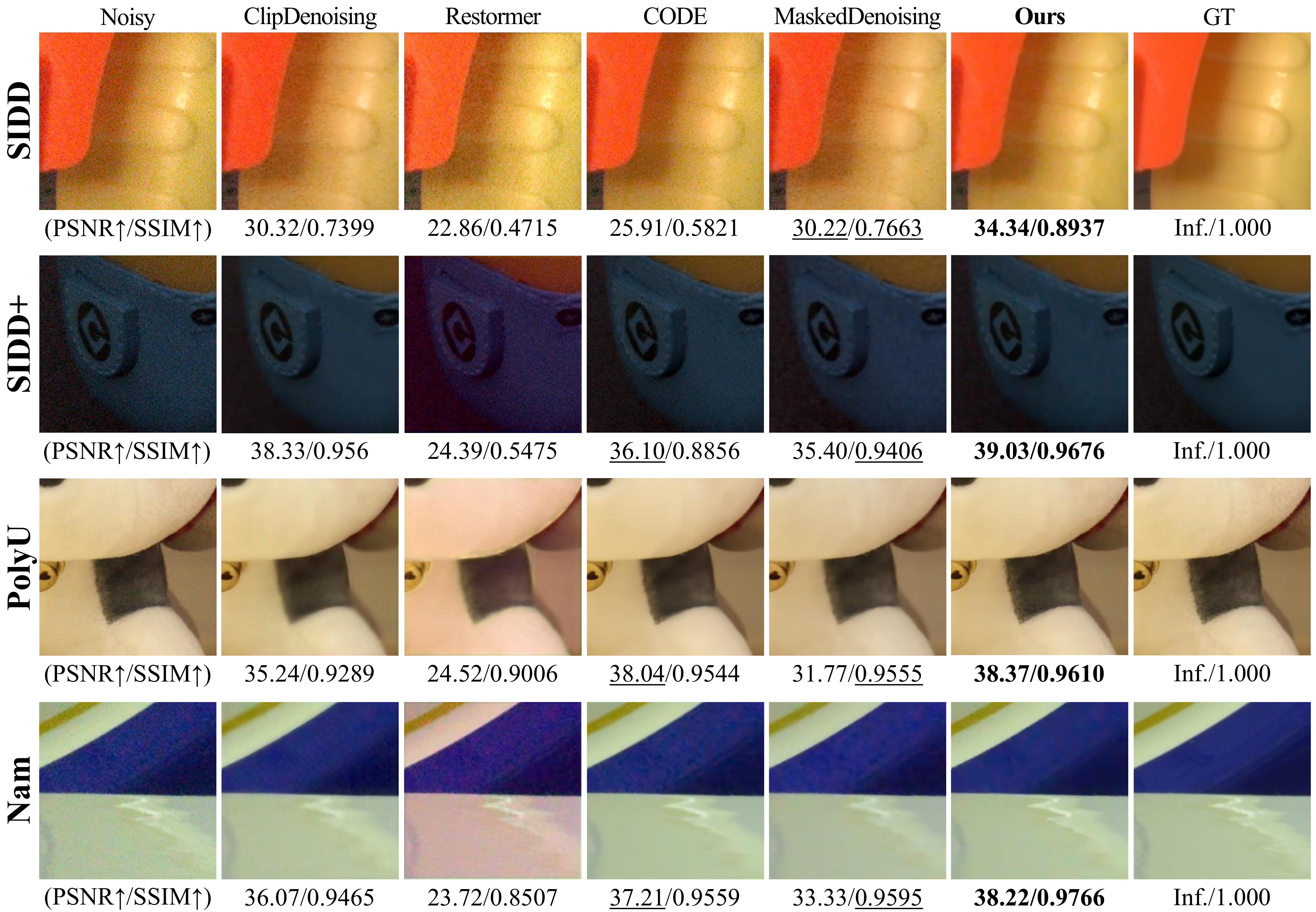}}
\caption{Qualitative results of denoising performance on real-world OOD noise in terms of PSNR$\uparrow$/SSIM$\uparrow$. 
Our method consistently produces the most visually natural output, with accurate color restoration and noise removal.}
\label{fig:real_world_denoising}
\vspace{-12mm}
\end{center}
\end{figure}

\noindent\textbf{Denoising Results on Synthetic Noise.}
To evaluate generalization performance under unseen conditions, we first evaluate \framework{} on diverse synthetic noise types across multiple intensity levels.
\ttabref{tab:synthetic_denoising} reports two variants of our method: \textbf{Ours$^\dagger$}, which applies the proposed DIC for adaptive denoising steps, and \textbf{Ours}, which performs denoising with fixed iterations ($T=10$).
The results show that both variants achieve SOTA performance on diverse OOD noise configurations. 
\ttabref{tab:abl_synthetic_denoising_extension} in the Supplementary Material provides additional in-distribution scores and extended results.

By contrast, self-attention networks such as SwinIR and Restormer tend to overfit to the specific noise patterns present in their training data due to their high capacity, which limits their effectiveness in denoising tasks that require broad generalization.
Moreover, lightweight or efficient networks like DnCNN and CODE also show inferior generalization performance, demonstrating that simply decreasing the number of parameters or focusing on efficiency does not adequately improve generalization performance.

The full model (Ours) shows the best results across all settings except salt-and-pepper noise removal, where it ranks second.
Our DIC-enabled variant (Ours$^\dagger$) achieves comparable performance while reducing the number of iterations by about 30$\%$ compared to the full model.

\begin{table*}[t]
  \centering
  \caption{Quantitative results of denoising performance on SIDD validation, SIDD+, PolyU, and Nam in terms of PSNR$\uparrow$ and SSIM$\uparrow$. All methods are trained with Gaussian noise with level of $\sigma=15$. The symbol $\dagger$ denotes that our model utilizes the proposed DIC during inference. The best and second-best results are highlighted in \textbf{bold} and \underline{underline}.}
  \vspace{-2.5mm}
  \resizebox{0.625\textwidth}{!}{ % 0.48
    \begin{tabular}{ccccccccccc|cc}
    \toprule
    \multicolumn{1}{r}{\multirow{2}{*}{Datasets}} & \multicolumn{2}{|c|}{ClipDenoising} & \multicolumn{2}{c}{Restormer} & \multicolumn{2}{c}{CODE} & \multicolumn{2}{c}{MaskedDenoising} & \multicolumn{2}{c}{\textbf{\textbf{Ours}}$^\dagger$} & \multicolumn{2}{c}{\textbf{Ours}} \\
\cmidrule(lr){2-3} \cmidrule(lr){4-5} \cmidrule(lr){6-7} \cmidrule(lr){8-9} \cmidrule(lr){10-11} \cmidrule(lr){12-13}          & \multicolumn{2}{|c|}{PSNR$\uparrow$/SSIM$\uparrow$}  & \multicolumn{2}{c}{PSNR$\uparrow$/SSIM$\uparrow$}  & \multicolumn{2}{c}{PSNR$\uparrow$/SSIM$\uparrow$}  & \multicolumn{2}{c}{PSNR$\uparrow$/SSIM$\uparrow$}  & \multicolumn{2}{c}{PSNR$\uparrow$/SSIM$\uparrow$}  & \multicolumn{2}{c}{PSNR$\uparrow$/SSIM$\uparrow$} \\
    \midrule
    SIDD   & \multicolumn{2}{|c|}{30.19/0.6409} & \multicolumn{2}{c}{22.54/0.3700} & \multicolumn{2}{c}{26.71/0.5149} & \multicolumn{2}{c}{28.65/0.6043} & \multicolumn{2}{c}{\underline{30.99}/\underline{0.7221}} & \multicolumn{2}{c}{\textbf{32.08}/\textbf{0.7578}} \\
    SIDD+  & \multicolumn{2}{|c|}{32.04/0.7340} & \multicolumn{2}{c}{24.45/0.4700} & \multicolumn{2}{c}{31.08/0.6772} & \multicolumn{2}{c}{31.52/0.7248} & \multicolumn{2}{c}{\underline{33.02}/\underline{0.7872}} & \multicolumn{2}{c}{\textbf{33.72}/\textbf{0.8117}} \\
    PolyU  & \multicolumn{2}{|c|}{34.83/0.9009} & \multicolumn{2}{c}{27.27/0.8309} & \multicolumn{2}{c}{\textbf{38.11}/0.9504} & \multicolumn{2}{c}{34.65/0.9333} & \multicolumn{2}{c}{\underline{38.05}/\textbf{0.9617}} & \multicolumn{2}{c}{37.93/\underline{0.9601}} \\
    Nam    & \multicolumn{2}{|c|}{35.45/0.9276} & \multicolumn{2}{c}{27.71/0.8145} & \multicolumn{2}{c}{\textbf{39.78}/0.9539} & \multicolumn{2}{c}{34.80/0.9416} & \multicolumn{2}{c}{38.36/\underline{0.9596}} & \multicolumn{2}{c}{\underline{38.96}/\textbf{0.9678}} \\
    \bottomrule
    \end{tabular}%
    }
  \label{tab:real_world_denoising}%
\vspace{-4.5mm}
\end{table*}%

\noindent\textbf{Denoising Results on Real-World sRGB Noise.}
We further test \framework{} on real-world datasets acquired with both smartphone and DSLR cameras.
Unlike synthetic noise, real-world noise contains intricate characteristics such as spatial correlations introduced by the non‑linear processing in image signal processing (ISP), providing a more challenging OOD denoising benchmark.

\ttabref{tab:real_world_denoising} summarizes the performance of our framework on four challenging real-world datasets: the SIDD validation set, SIDD+, PolyU, and Nam.
Both the full model (Ours) and its DIC-enabled variant (Ours$^{\dagger}$) achieve superior performance on most datasets, consistently outperforming other methods. 
While CODE achieves slightly higher scores on PolyU and Nam, it shows inferior results on SIDD and SIDD+, indicating possible overfitting to a specific domain.

Overall, our experiments demonstrate that the proposed denoising framework significantly improves both numerical and perceptual quality over SOTA methods in real-world noisy images. 
This robust performance across diverse datasets highlights the effectiveness and generalization capability of our approach in real-world denoising scenarios.

\begin{figure}
\begin{center}
\centerline{\includegraphics[width=1\columnwidth]{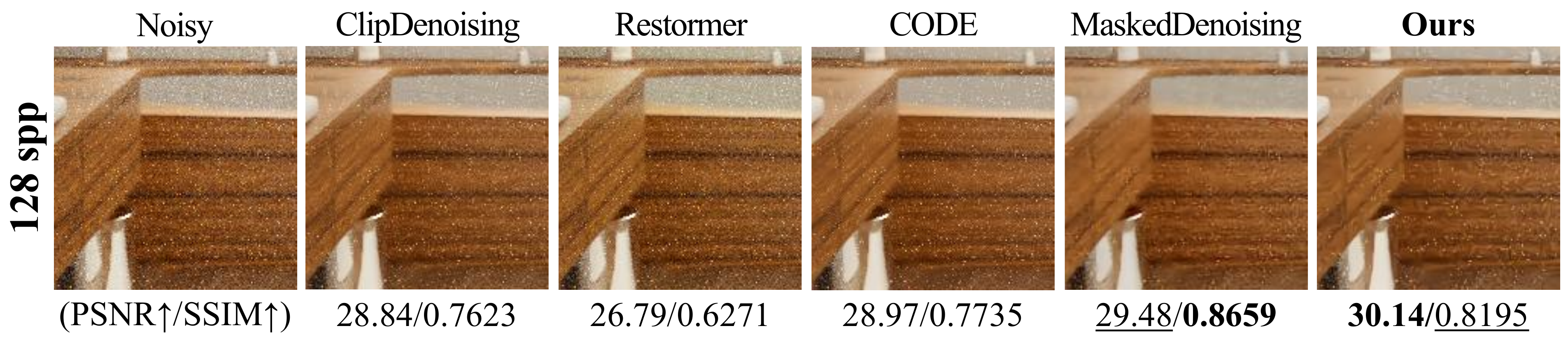}}
\caption{Qualitative results of denoising performance on Monte Carlo rendering OOD noise in terms of PSNR$\uparrow$/SSIM$\uparrow$. Our approach better preserves surface details and geometry boundaries while effectively removing stochastic noise.}
\label{fig:monte_carlo_denoising}
\vspace{-12mm}
\end{center}
\end{figure}

\noindent\textbf{Denoising Results on Monte Carlo Rendering Noise.}
We further evaluate our method on Monte Carlo-rendered images, where rendering algorithms simulate light transport to produce realistic images, but introduce noise due to the inherent randomness in sampling when approximating the rendering equation~\cite{mc_algo_1,mc_algo_2,mc_algo_3}.
In particular, noise is generated due to the limited number of samples per pixel (spp) during the rendering process, where lower sample rates lead to more severe noise artifacts.
\ttabref{tab:monte_carlo_denoising} summarizes the quantitative denoising results for two sampling rates: 64 and 128 spp.

For images rendered with 64 spp, our full model outperforms competing methods such as CODE, MaskedDenoising, and even CLIPDenoising in terms of PSNR, while achieving competitive SSIM scores.
Notably, the DIC-enabled variant (Ours$^\dagger$) also demonstrates robust performance.
At the higher sampling rate with 128 spp, our method further enhances image quality, consistently showing the best PSNR and the second-highest SSIM.
Overall, the findings confirm that our denoising framework effectively removes Monte Carlo noise artifacts, even when the renderings are generated with a small number of spp (\iie under high-noise conditions).

\begin{table}[t]
  \centering
  \caption{Quantitative results of denoising performance on Monte Carlo-rendered image. The symbol $\dagger$ denotes that our model utilizes the proposed DIC during inference. The best and second-best results are highlighted in \noindent\textbf{bold} and \underline{underline}.}
  \vspace{-1.5mm}
  \resizebox{0.35\textwidth}{!}{
    \begin{tabular}{c|cccc}
    \toprule
    \multicolumn{1}{c}{\multirow{2}{*}{Method}} & \multicolumn{2}{|c}{64 samples per pixel} & \multicolumn{2}{|c}{128 samples per pixel} \\
\cmidrule(lr){2-3} \cmidrule(lr){4-5}          & \multicolumn{2}{c|}{PSNR$\uparrow$/SSIM$\uparrow$}  & \multicolumn{2}{c}{PSNR$\uparrow$/SSIM$\uparrow$} \\
    \midrule
    ClipDenoising  & \multicolumn{2}{c}{27.36/0.7214} & \multicolumn{2}{|c}{30.52/0.8070} \\ \midrule
    Restormer  & \multicolumn{2}{c}{24.33/0.6123} & \multicolumn{2}{|c}{26.45/0.6896} \\
    CODE   & \multicolumn{2}{c}{27.04/0.7109} & \multicolumn{2}{|c}{30.52/0.8020} \\
    MaskedDenoising  & \multicolumn{2}{c}{\underline{27.94}/\textbf{0.7734}} & \multicolumn{2}{|c}{30.29/\textbf{0.8477}} \\
    \midrule
    \textbf{Ours}$^\dagger$   & \multicolumn{2}{c}{27.83/0.7286} & \multicolumn{2}{|c}{\underline{30.98}/0.8128} \\
    \textbf{Ours}   & \multicolumn{2}{c}{\textbf{28.09}/\underline{0.7446}} & \multicolumn{2}{|c}{\textbf{31.13}/\underline{0.8259}} \\
    \bottomrule
    \end{tabular}%
    }
  \label{tab:monte_carlo_denoising}%
\vspace{-2.5mm}
\end{table}%

\subsection{Ablation Studies}
Please refer to the Supplementary Material~\ssecref{sec:supp_additional_abl} for additional ablation studies.

\noindent\textbf{Model Size and Inference Time.}
\begin{table}[t]
\centering
\caption{Comparison of model efficiency in terms of the number of parameters, inference time, and FLOPs, with all methods evaluated on images of size $3 \times 160 \times 160$. $\dagger$ denotes that our method was measured in a DIC environment. The best, second-lightest, and fastest results are highlighted in \noindent\textbf{bold} and \underline{underlined}.}
\vspace{-1.5mm}
\resizebox{0.475\textwidth}{!}{
\begin{tabular}{c|c|cccccc}
\toprule
Model                 & \makecell{CLIP \\ Denoising} & DnCNN & Restormer & CODE  & \makecell{Masked \\ Denoising} & \textbf{Ours}$^\dagger$ & \textbf{Ours} \\ \midrule
Params. (M)            & 19.55 & \underline{0.67} & 26.1 & 12.2  & 0.82 & \textbf{0.04}     & \textbf{0.04} \\
Infer. Time(s)         & 0.004 & \textbf{0.002}   & 0.03 & 0.05  & 0.14 & \underline{0.004} & 0.005 \\
FLOPs (G)             & 8.77 & 17.2  & 60.5 & 19.6  & 24.9 & \textbf{8.02}     & \underline{11.5} \\ \hline
\end{tabular}
}
\label{tab:model_size_and_inference_time}%
\vspace{-4.5mm}
\end{table}
\ttabref{tab:model_size_and_inference_time} summarizes the model size, inference time, and floating-point operations (FLOPs).
Notably, our method has only 0.04 million parameters, making it approximately 17 times smaller than DnCNN, the second smallest model. 
Moreover, the DIC-enabled variant (Ours$^\dagger$) exhibits the fewest FLOPs, requiring only about half the computational budget of DnCNN, which highlights the efficiency of its lightweight architecture.
Although our method has the smallest number of parameters and FLOPs, its inference speed is slower than that of DnCNN due to suboptimal operations. 
Specifically, optimizing the inefficient PyTorch unfolding operation for patch extraction will significantly enhance the inference speed.
In summary, \framework{} shows robust performance for OOD noise removal while providing significant efficiency benefits.

\noindent\textbf{Global Statistics and Local Correlation Module.}
We evaluate the contributions of the GSM and LCM via the ablation study in \ttabref{tab:abl_global_local}, where model variants were trained to investigate the effect of each module individually.
Integrating local structural cues through correlation maps with global statistics extracted from image residuals significantly improves image quality across diverse noise conditions.

\begin{table}[!t]
    \centering
    \begin{minipage}{.48\linewidth}

    \centering
  \caption{Effect of GSM and LCM. The best result is highlighted in \noindent\textbf{bold}.}
  \vspace{-1.5mm}
  \resizebox{1.0\textwidth}{!}{
    \begin{tabular}{cccc}
    \toprule
    \multirow{2}{*}{GSM} & \multirow{2}{*}{LCM} & Mixture & Spatial Gaussian \\ \cmidrule(lr){3-4} 
                         &                      & PSNR$\uparrow$/SSIM$\uparrow$     & PSNR$\uparrow$/SSIM$\uparrow$        \\ \midrule[0.2pt]
    \xm                    & \xm                    & 27.07/0.8225  & 27.70/0.8248     \\
    \cm                    & \xm                    & 27.33/0.8330  & 27.44/0.8159     \\
    \xm                    & \cm                    & 27.25/0.8283  & 27.76/0.8282     \\
    \cm                    & \cm                    & \textbf{27.52}/\textbf{0.8405}  & \textbf{27.78}/\textbf{0.8333}      \\
    \bottomrule
    \end{tabular}%
    }
  \label{tab:abl_global_local}%
    
    \end{minipage}
    \hspace{0.1cm}
    \begin{minipage}{.48\linewidth}

\centering
  \caption{Effect each component in DID block. The best result is highlighted in \noindent\textbf{bold}.}
  \vspace{-1.5mm}
  \resizebox{1.0\textwidth}{!}{
    \begin{tabular}{lcc}
    \toprule
    \multirow{2}{*}{Methods} & Mixture & Spatial Gaussian \\ \cmidrule(lr){2-3} 
                  & PSNR$\uparrow$/SSIM$\uparrow$     & PSNR$\uparrow$/SSIM$\uparrow$              \\ \midrule[0.2pt]
    \textbf{Baseline (Full)}          & \textbf{27.52}/\textbf{0.8405}  & \textbf{27.78}/\textbf{0.8333}           \\ \midrule 
    (\xm) RMS Norm.       & 27.13/0.8267  & 27.55/0.8201           \\
    (\xm) Dilation         & 27.48/0.8352  & 27.37/0.8121           \\
    (\xm) Power Norm.       & 23.02/0.7435  & 24.04/0.7485           \\
    \bottomrule
    \end{tabular}%
    }
  \label{tab:abl_did_components}%
    
    \end{minipage}
    \vspace{-10px}
\end{table}

%%%%%%%%%%%%%%%%%%%%%

\noindent\textbf{Impact of DID Components.}
\ttabref{tab:abl_did_components} summarizes an ablation study on the three key components of the DID block: RMS Normalization, unfolding dilation, and Power Normalization.
Model variants are trained with each component removed.
The results indicate that every component independently enhances the model’s robustness, consistently improving performance on diverse OOD noise types.

\section{Conclusion}
In this work, we introduced \framework{}, a novel image denoising framework that integrates dynamic kernel prediction with an adaptive iterative refinement strategy. 
Our method extracts noise-invariant features using the shallow Feature Extraction Module (FEM), enriched by the Global Statistics Module (GSM) and Local Correlation Module (LCM).
Then the Kernel Prediction Module (KPM) generates per-pixel kernels for content-aware filtering using these features.
The Dynamic Iteration Control (DIC) adaptively adjusts the number of denoising iterations according to kernel convergence, thereby reducing computational cost and preventing over-smoothing in low-noise regions. This approach ensures consistently high-quality restoration across diverse noise intensities.
Although \framework{} was trained exclusively on Gaussian noise at one specific level, it exhibits strong performance with various noise types and retains an impressively small model size ($\sim$ 0.04M parameters).
Experimental results confirm that the use of dynamic kernel prediction combined with iterative refinement successfully leads to robust generalization across different domains.

\clearpage
\noindent
\section*{Acknowledgments}
This work was supported by National Research Foundation of Korea (NRF) grant funded by the Korea government (MSIT) (RS-2023-00222776), and Institute of Information communications Technology Planning Evaluation (IITP) grant funded by the Korea government (MSIT) (No.2022- 0-00156, Fundamental research on continual meta-learning for quality enhancement of casual videos and their 3D metaverse transformation) and Institute of Information \& communications Technology Planning \& Evaluation (IITP) grant funded by the Korea government(MSIT) (No.RS-2020-II201373, Artificial Intelligence Graduate School Program(Hanyang University)) and the research fund of Hanyang University(HY-2025).
{
    \small
    \bibliographystyle{ieeenat_fullname}
    \bibliography{main}
}

\clearpage
\setcounter{page}{1}
\maketitlesupplementary

\begingroup
  \setcounter{figure}{0}
  \setcounter{table}{0}
  \setcounter{equation}{0}
  \setcounter{section}{0}
  \setcounter{algocf}{0}

  \renewcommand{\thesection}{A\arabic{section}}
  \renewcommand{\thefigure}{A\arabic{figure}}
  \renewcommand{\thetable}{A\arabic{table}}
  \renewcommand{\theequation}{AA\arabic{equation}}
  \renewcommand{\thealgocf}{A\arabic{algocf}}

\section{Appendix}
\label{sec:appendix}

\subsection{Details Regarding Test Noise}
\label{appendix:noise_details}
To rigorously evaluate OOD denoising robustness, we benchmark eight noise categories:
(1) real-world noise captured by smartphone and DSLR cameras;
(2) Monte Carlo (MC)-rendered noise $spp \in \{64, 128\}$;
(3) additive Gaussian noise with $\sigma \in \{15,25,50\}$;
(4) spatially correlated Gaussian noise with $\sigma \in \{45,50,55\}$;
(5) Poisson noise with $\alpha \in \{2.5,3.0,3.5\}$;
(6) speckle noise with $\sigma \in \{0.02,0.03,0.04\}$;
(7) salt-and-pepper noise with $p \in \{0.012,0.016,0.02\}$; and
(8) mixture noise at levels $\{1,2,3,4\}$.
Following MaskedDenoising~\cite{masked_denoising}, Gaussian and spatial Gaussian noise levels are rescaled to $[0,255]$, whereas the remaining noise levels are normalized to $[0,1]$.

Further implementation details are provided in the subsequent subsections.

\noindent\textbf{Spatial Gaussian Noise.}  
Spatial Gaussian noise differs from standard Gaussian noise in that its values are spatially correlated rather than independent across pixels. This correlation often arises from sensor imperfections or smoothing effects during image processing. Following MaskedDenoising~\cite{masked_denoising}, we synthesize spatial Gaussian noise by convolving \textit{i.i.d.} Gaussian noise with standard deviation $\sigma$ using a $3 \times 3$ averaging filter.

\noindent\textbf{Poisson Noise.}  
Poisson noise, often referred to as photon or shot noise, originates from the quantized nature of light. Its variance is equal to the mean signal level, making it inherently signal-dependent. This type of noise is particularly prevalent in low-light conditions, where the photon count is low. We synthesize Poisson noise as follows:
$
\mathrm{\mathbf{I}}_{\mathrm{Noisy}} = \mathrm{\mathbf{I}}_{\mathrm{Clean}} + \mathrm{\mathbf{n}} \cdot \alpha,
$
where $\mathbf{n}$ is sampled from a Poisson distribution and $\alpha$ controls the noise magnitude.

\noindent\textbf{Salt-and-Pepper Noise.}  
Salt-and-pepper noise is an impulsive corruption characterized by random occurrences of extreme pixel intensities (\ie pure black or white). This artifact commonly stems from transmission errors or sensor defects. Following MaskedDenoising~\cite{masked_denoising}, we synthesize salt-and-pepper noise with MATLAB’s \texttt{imnoise} function.

\noindent\textbf{Speckle Noise.}  
Speckle noise is a multiplicative perturbation commonly observed in coherent imaging modalities such as synthetic aperture radar (SAR) and ultrasound; it originates from the interference of multiple scattered wavefronts and manifests as granular texture. Following MaskedDenoising~\cite{masked_denoising}, we synthesize speckle noise with MATLAB’s \texttt{imnoise} function.

\noindent\textbf{Mixture Noise.}  
The mixture noise model combines several noise sources, including Gaussian, Poisson, speckle, and salt-and-pepper noise, to emulate real-world noise characteristics. Following MaskedDenoising~\cite{masked_denoising}, we synthesize this mixture by sequentially adding Gaussian noise with variance $\sigma_g^{2}$, speckle noise with variance $\sigma_{s1}^{2}$, Poisson noise scaled by $\alpha$, salt-and-pepper noise with density $d$, and a second speckle component with variance $\sigma_{s2}^{2}$.

We categorize the mixture noise into four levels, determined by their overall intensity and complexity:

$
\textbf{Level 1:} \quad \sigma^2_g = 0.003,\quad \sigma^2_{s1} = 0.003,\quad \alpha = 1,\quad d = 0.002,\quad \sigma^2_{s2} = 0.003,
$

$
\textbf{Level 2:} \quad \sigma^2_g = 0.004,\quad \sigma^2_{s1} = 0.004,\quad \alpha = 1,\quad d = 0.002,\quad \sigma^2_{s2} = 0.003,
$

$
\textbf{Level 3:} \quad \sigma^2_g = 0.006,\quad \sigma^2_{s1} = 0.006,\quad \alpha = 1,\quad d = 0.003,\quad \sigma^2_{s2} = 0.006,
$

$
\textbf{Level 4:} \quad \sigma^2_g = 0.008,\quad \sigma^2_{s1} = 0.008,\quad \alpha = 1,\quad d = 0.004,\quad \sigma^2_{s2} = 0.008.
$

Note that, for each level, the individual noise components are introduced in the prescribed sequence.

\noindent\textbf{Monte Carlo-Rendered Image Noise.}  
Noise produced by the stochastic Monte Carlo integration underlying physically based rendering results in sampling artifacts that reveal the finite number of rays used to approximate light transport. We evaluate the proposed method on the Monte Carlo noise benchmark from~\cite{mc_dataset}.

\noindent\textbf{Real-World Noise.}  
Real-world camera images contain complex noise arising from photon statistics, sensor readout, and in-camera ISP post-processing. To evaluate the robustness of \framework{} in these conditions, we evaluate it on diverse datasets captured with both smartphone and DSLR devices~\cite{sidd, siddplus, polyu, nam}.

\begin{table}[t]
\centering
    \caption{Comparison of denoising performance based on model capacity. The number of parameters for DnCNN and Restormer are set to be similar to those of our method. The best results are highlighted in \textbf{bold}.}
    \resizebox{0.495\textwidth}{!}{
\begin{tabular}{ccccc}
\toprule[0.5pt]
\multirow{2}{*}{Methods} & \multirow{2}{*}{\makecell{Params. \\ (M)}} & Gaussian & Spatial Gaussian & SIDD \\ \cmidrule(lr){3-5} 
                         & & PSNR$\uparrow$/SSIM$\uparrow$        & PSNR$\uparrow$/SSIM$\uparrow$    & PSNR$\uparrow$/SSIM$\uparrow$    \\ \midrule[0.2pt]
DnCNN                    & 0.04 & 19.74/0.4262     & 27.24/0.7966 & 28.93/0.6040 \\
Restormer                & 0.04 & 20.11/0.4389     & 26.39/0.7572 & 27.39/0.5388 \\ \midrule[0.2pt]
\textbf{Ours}            & 0.04 & \textbf{25.06}/\textbf{0.7547}  & \textbf{27.78}/\textbf{0.8333} & \textbf{32.08}/\textbf{0.7578} \\ \bottomrule[0.5pt]
\end{tabular}
\label{tab:abl_model_capacity_comparison}
}
\vspace{-2.5mm}
\end{table}

\begin{table*}[!ht]
\centering
\vspace{-2.5mm}
\caption{OOD denoising results on various training noise types. The best results are highlighted in \textbf{bold}.}
\vspace{-2.5mm}
\resizebox{0.75\textwidth}{!}{\begin{tabular}{c|c|ccc|c}
\toprule
\multicolumn{1}{c|}{\multirow{3}{*}{\begin{tabular}[c]{@{}c@{}}Training \\ Noise Type\end{tabular}}} & \multirow{3}{*}{Model} & \multicolumn{1}{c}{\multirow{2}{*}{\begin{tabular}[c]{@{}c@{}}Real-World \\ (SIDD)\end{tabular}}} & \multicolumn{1}{c}{\multirow{2}{*}{\begin{tabular}[c]{@{}c@{}}Spatial Gaussian\\ ($\sigma=55$)\end{tabular}}} & \multicolumn{1}{c}{\multirow{2}{*}{\begin{tabular}[c]{@{}c@{}}Medical Imaging\\ (LDCT)\end{tabular}}} & \multicolumn{1}{|c}{\multirow{2}{*}{Average}} \\
\multicolumn{1}{c|}{}                                     &                         & \multicolumn{1}{c}{}                                                                                   & \multicolumn{1}{c}{}                                  & \multicolumn{1}{c}{}                                  & \multicolumn{1}{|c}{}                         \\ \cmidrule(lr){3-6}
\multicolumn{1}{c|}{}                                     &                         & \multicolumn{1}{c}{PSNR$\uparrow$/SSIM$\uparrow$}                                                                        & \multicolumn{1}{c}{PSNR$\uparrow$/SSIM$\uparrow$}                       & \multicolumn{1}{c}{PSNR$\uparrow$/SSIM$\uparrow$}                       & \multicolumn{1}{|c}{PSNR$\uparrow$/SSIM$\uparrow$}              \\ \hline
            \multirow{3}{*}{\begin{tabular}[c]{@{}c@{}}Gauss. $\sigma=15$ \\ (Baseline)\end{tabular}}                                           & CGNet             & 27.32/0.5359          & 25.41/0.6651          & 38.50/0.7878          & 30.41/0.6629          \\
                                                                                                                                                & MaskedDenoising                & 28.65/0.6043          & 26.72/0.7685          & 37.76/0.7880          & 31.04/0.7203          \\
                                                                                                                                                & \textbf{Ours}     & \textbf{32.08/0.7578} & \textbf{27.87/0.8049} & \textbf{44.45/0.9643} & \textbf{34.80/0.8423} \\ \hline
\multirow{3}{*}{\begin{tabular}[c]{@{}l@{}}(i) Gauss. $\sigma\sim\mathcal{U}(15,50)$\end{tabular}}                                              & CGNet             & 25.36/0.4298          & 25.09/0.6492          & 43.69/0.9254          & 31.38/0.6681          \\
                                                                                                                                                & MaskedDenoising                & 27.93/0.7320          & 25.07/0.7438          & 27.96/0.6870          & 26.99/0.7209          \\
                                                                                                                                                & \textbf{Ours}     & \textbf{32.39/0.7786} & \textbf{28.08/0.8056} & \textbf{44.88/0.9684} & \textbf{35.12/0.8509} \\ \hline
\multirow{3}{*}{\begin{tabular}[c]{@{}c@{}}(ii) Gauss. $\sigma\sim\mathcal{U}(15,50)$\\ + \\ Poisson $\alpha\sim\mathcal{U}(1,4)$\end{tabular}} & CGNet             & 25.84/0.4440          & 25.48/0.6683          & 41.46/0.8737          & 30.93/0.6620          \\
                                                                                                                                                & MaskedDenoising                & 27.99/0.7433          & 25.03/0.7427          & 24.52/0.5447          & 25.85/0.6769          \\
                                                                                                                                                & \textbf{Ours}     & \textbf{32.95/0.8197} & \textbf{27.95/0.8004} & \textbf{44.60/0.9632} & \textbf{35.17/0.8611} \\ \hline
\multirow{3}{*}{\begin{tabular}[c]{@{}l@{}}(iii) Monte Carlo 16 spp\end{tabular}}                                                               & CGNet             & 27.31/0.5227          & 23.85/0.5861          & \textbf{44.70/0.9691} & 31.95/0.6926          \\
                                                                                                                                                & MaskedDenoising                & 31.45/0.7359          & 25.24/0.6657          & 44.49/0.9674          & 33.73/0.7897          \\
                                                                                                                                                & \textbf{Ours}     & \textbf{34.13/0.8414} & \textbf{26.10/0.7170} & 44.51/0.9625          & \textbf{34.91/0.8403} \\ \bottomrule
\end{tabular}
}
\label{tab:training_noise_condition_denoising_result}%
\vspace{-4mm}
\end{table*}

\subsection{Additional Ablation Studies}
\label{sec:supp_additional_abl}
\noindent\textbf{Impact of Model Capacity on Generalization.}
To investigate whether reducing the number of model parameters alone can mitigate overfitting and enhance generalization, we use two conventional denoisers, DnCNN~\cite{dncnn} and Restormer~\cite{restormer}, with capacities adjusted to match our model. These variants are evaluated on Urban100~\cite{urban} corrupted with Gaussian noise ($\sigma = 50$) and spatial Gaussian noise ($\sigma = 55$), as well as on the SIDD real-world dataset. As reported in \ttabref{tab:abl_model_capacity_comparison}, simply reducing network size does not meaningfully alleviate overfitting because noise characteristics vary substantially across types and levels. In contrast, our framework achieves the best performance across all noise settings while remaining the most compact, thanks to its modules that dynamically adapt to OOD noise.

\noindent\textbf{Impact of Diverse Training Noise Conditions on Generalization.}
We perform ablation studies to evaluate how well our method generalizes under varying training noise conditions.
\ttabref{tab:training_noise_condition_denoising_result} summarizes the performance of IDF, MaskedDenoising~\cite{masked_denoising}, and the recent SOTA model CGNet~\cite{cgnet}, each trained under three distinct noise regimes (i–iii).
All models are evaluated on challenging OOD noise scenarios: real-world sensor noise from SIDD \cite{sidd}, spatially correlated Gaussian noise on CBSD68 \cite{cbsd}, and low-dose CT (LDCT) scans \cite{low_dose_ct}.
The LDCT benchmark, whose intensity distribution differs markedly from natural sRGB images, serves as an extreme test of OOD robustness.
Because LDCT volumes are single-channel, each slice is replicated across three RGB channels to avoid architectural changes.
Despite CGNet’s 119M parameters and MaskedDenoising’s mixed-noise curriculum, both methods suffer substantial performance drops in these settings.
In contrast, IDF consistently preserves high fidelity across all benchmarks, and its accuracy improves as the training-noise configuration becomes more heterogeneous, highlighting its noise-invariant filtering capability.
Please note that, for the LDCT evaluation, we constrain IDF to a single denoising iteration to avoid over-smoothing while maintaining computational efficiency.

\begin{table}[t]
\centering
    \caption{Comparison results on denoising performance depending on different kernel size $K$ in~\eeqref{eq:patch_conv}.
    The best results are highlighted in \textbf{bold}.}
    \resizebox{0.375\textwidth}{!}{
\begin{tabular}{ccc}
\toprule[0.5pt]
\multirow{2}{*}{\makecell{Kernel Size \\ ($K$)}} & Mixture & Spatial Gaussian \\ \cmidrule(lr){2-3} 
                  & PSNR$\uparrow$/SSIM$\uparrow$     & PSNR$\uparrow$/SSIM$\uparrow$              \\ \midrule[0.2pt]
\textbf{3}        & \textbf{27.52}/\textbf{0.8405}  & \textbf{27.78}/\textbf{0.8333}           \\
5                 & 27.27/0.8366  & 27.58/0.8259           \\
7                 & 27.30/0.8356  & 27.41/0.8143           \\ \bottomrule[0.5pt]
\end{tabular}
    \label{tab:abl_kernel_size}
    }
\end{table}

\noindent\textbf{Impact of Kernel Size.}
\ttabref{tab:abl_kernel_size} analyzes how the patch-convolution kernel size $K$ (see \eeqref{eq:patch_conv}) affects denoising performance on mixture and spatial Gaussian noise. A kernel of $K = 3$ shows the highest PSNR and SSIM on both noise types. Enlarging the kernel to $K = 5$ or $K = 7$ slightly degrades performance, indicating that small receptive fields capture essential local details without introducing excessive smoothing. These results show that a $3 \times 3$ kernel offers the best balance between detail preservation and noise removal within our framework.

\begin{table}[t]
\centering
    \caption{Comparison results on denoising performance depending on different power normalization factor $p$ in~\eeqref{eq:power_norm}.
    The best results are highlighted in \textbf{bold}.}
    \resizebox{0.4\textwidth}{!}{
\begin{tabular}{ccc}
\toprule[0.5pt]
\multirow{2}{*}{\makecell{Power Norm. \\ ($p$)}} & Mixture & Spatial Gaussian \\ \cmidrule(lr){2-3} 
                  & PSNR$\uparrow$/SSIM$\uparrow$     & PSNR$\uparrow$/SSIM$\uparrow$              \\ \midrule[0.2pt]
1          & 27.01/0.8289  & 27.34/0.8231           \\
2          & 27.24/0.8325  & 27.65/0.8293           \\
\textbf{3} & \textbf{27.52}/\textbf{0.8405}  & \textbf{27.78}/\textbf{0.8333}           \\
4          & 27.50/0.8398  & 27.55/0.8199           \\ \bottomrule[0.5pt]
\end{tabular}
    \label{tab:power_norm_factor}
    }
\end{table}

\noindent\textbf{Impact of the Power Normalization Factor.}
We perform an ablation study to investigate the impact of the power normalization factor $p$ (see~\eeqref{eq:power_norm}) on denoising performance. The results are presented in~\ttabref{tab:power_norm_factor} for two noise types: mixture noise and spatial Gaussian noise. Four different normalization factors, ranging from 1 to 4, are evaluated.

For mixture noise, performance consistently improves as the normalization factor increases from $p=1$ to $p=3$, with the highest PSNR observed at $p=3$. A similar trend is noted for spatial Gaussian noise, where $p=3$ also shows the best results among the evaluated settings. Increasing the normalization factor to $p=4$ does not lead to further improvements; instead, it appears to reduce kernel diversity, which negatively impacts performance. These results indicate that a power normalization factor of $p=3$ provides the most effective denoising performance across both noise types.

\begin{table*}[t]
\centering
    \caption{Comparison results on denoising performance depending on different DIC threshold $\kappa$ in~\eeqref{eq:stop_threshold}. 
    We choose $\kappa=0.015$ for main results and corresponding scores are highlighted in \textbf{bold}.}
    \resizebox{1.0\textwidth}{!}{
\begin{tabular}{c|cc|cc|cc|cc}
\toprule[0.5pt]
\multirow{3}{*}{\makecell{Threshold \\ ($\kappa$)}} & \multicolumn{8}{c}{Spatial Gaussian}                                                                                      \\ \cmidrule(lr){2-9} 
                           & \multicolumn{2}{c}{$\sigma=45$} & \multicolumn{2}{|c}{$\sigma=50$} & \multicolumn{2}{|c}{$\sigma=55$} & \multicolumn{2}{|c}{Average}  \\ \cmidrule(lr){2-9} 
                           & PSNR$\uparrow$/SSIM$\uparrow$    & \# Iterations & PSNR$\uparrow$/SSIM$\uparrow$    & \# Iterations & PSNR$\uparrow$/SSIM$\uparrow$    & \# Iterations & PSNR$\uparrow$/SSIM$\uparrow$    & \# Iterations \\ \midrule[0.2pt]
% 0                          & 28.98/0.8619 & 10.00         & 28.37/0.8483 & 10            & 27.78/0.8333 & 10.00         & 28.38/0.8478 & 10.00         \\
0.005                      & 28.89/0.8557 & 8.02          & 28.25/0.8432 & 8.22          & 27.70/0.8298  & 8.58          & 28.28/0.8429 & 8.27          \\
0.01                       & 28.80/0.8524 & 7.40          & 28.15/0.8383 & 7.54          & 27.65/0.8275 & 8.04          & 28.20/0.8394 & 7.66          \\
\textbf{0.015}                      & \textbf{28.77/0.8511} & \textbf{7.10}          & \textbf{28.12/0.8364} & \textbf{7.22}          & \textbf{27.54/0.8217} & \textbf{7.44}          & \textbf{28.14/0.8364} & \textbf{7.25}          \\
0.02                       & 28.73/0.8497 & 6.60          & 28.10/0.8340  & 6.72          & 27.49/0.8182 & 6.86          & 28.11/0.8340 & 6.73          \\
0.025                      & 28.69/0.8472 & 6.00          & 28.06/0.8324 & 6.32          & 27.43/0.8154 & 6.50          & 28.06/0.8317 & 6.27          \\
0.03                       & 28.65/0.8455 & 5.66          & 27.99/0.8290  & 5.78          & 27.42/0.8152 & 6.34          & 28.02/0.8299 & 5.93          \\ \bottomrule[0.5pt]
\end{tabular}
    \label{tab:iteration_threshold}
    }
\end{table*}

\noindent\textbf{Impact of DIC threshold.}
\ttabref{tab:iteration_threshold} presents an ablation study on the influence of the DIC threshold $\kappa$ (as defined in~\eeqref{eq:stop_threshold}) on denoising performance under spatial Gaussian noise at varying noise levels ($\sigma=\{45,50,55\}$). The results include PSNR, SSIM, and the average number of iterations before termination.

As $\kappa$ increases from 0.005 to 0.03, the average number of iterations decreases from approximately 8.27 to 5.93. This pattern is consistent across all noise levels, suggesting that a looser threshold prompts earlier termination of the iterative process, thereby reducing computational demands.

The findings reveal a clear trade-off: lower thresholds yield more iterations and slightly better denoising performance, whereas higher thresholds reduce computational cost at the expense of minor performance degradation. Based on this observation, we adopt $\kappa=0.015$ as the default setting, as it offers a balanced compromise between performance and efficiency.

The selection of $\kappa$ within our DIC mechanism directly affects both computational efficiency and denoising quality. For scenarios where computational constraints are important, a higher threshold may be preferable despite a modest performance loss. In contrast, for applications prioritizing maximum denoising fidelity, a lower $\kappa$ value that allows additional iterations may be more suitable. This flexibility allows the framework to adapt to diverse practical requirements.

\begin{table}[t]
\centering
    \caption{Comparison results on denoising performance depending on the different number of training full iterations $T$ in~\ffigref{fig:adaptive_iterative_scheme}.
    The best results are highlighted in \textbf{bold}.}
    \resizebox{0.375\textwidth}{!}{
\begin{tabular}{ccc}
\toprule[0.5pt]
\multirow{2}{*}{\makecell{\# Iterations \\ ($T$)}} & Mixture & Spatial Gaussian \\ \cmidrule(lr){2-3} 
                  & PSNR$\uparrow$/SSIM$\uparrow$     & PSNR$\uparrow$/SSIM$\uparrow$              \\ \midrule[0.2pt]
6          & 27.32/0.8326  & 27.48/0.8162           \\
8       & 27.37/0.8390  & 27.67/0.8287           \\
\textbf{10}         & \textbf{27.52}/\textbf{0.8405}  & 27.78/0.8333           \\
12       & 27.42/0.8389  & \textbf{27.86}/\textbf{0.8383}           \\ \bottomrule[0.5pt]
\end{tabular}
    \label{tab:abl_training_iteration}
    }
\end{table}

\noindent\textbf{Impact of the Total Number of DID Block Full Iteration.}
\ttabref{tab:abl_training_iteration} presents an ablation study on the impact of the number of full training iterations ($T$) on denoising performance for two noise types: mixture and spatial Gaussian. As shown in the table, varying the number of iterations leads to differences in PSNR and SSIM.

Denoising performance progressively improves as $T$ increases from 6 to 10, reaching its peak at $T=10$. A further increase to $T=12$ results in a slight performance decline, indicating that excessive iterations may cause over-smoothing. Considering both noise types and the trade-off between performance and computational cost, we select $T=10$ as the default setting. This choice strikes a balance between effective noise removal and detail preservation, while also highlighting the advantages of our DIC strategy in adapting the iterative denoising process based on image content and noise characteristics. These findings demonstrate the importance of selecting an appropriate $T$ to achieve optimal performance across diverse noise scenarios without incurring unnecessary computational overhead.

\noindent\textbf{Further Analysis on Inference Speed.}
In addition to the results presented in~\ttabref{tab:model_size_and_inference_time} of the main paper, we further evaluate the effectiveness of the proposed DIC in terms of computational efficiency. Specifically, we measure inference speed using high-resolution 4K images to evaluate whether \framework{} can be efficiently deployed in real-world scenarios.

While the runtime difference between the kernel-based DIC variant and the full-iteration version is minimal for $160\times160$ images, the benefit of DIC becomes significantly more evident at higher resolutions. On a single 4K frame, the DIC variant achieves a substantial 30\% speed-up (0.383s vs. 0.548s).

These results highlight that \framework{} offers strong performance in both OOD denoising and inference efficiency, making it well-suited for practical deployment.

\begin{algorithm}[t]
    \small
    \DontPrintSemicolon
    \SetAlgoLined
    \SetKwInOut{Require}{Require}
    \Require{Input noisy image $\mathbf{I}_\mathrm{Noisy}$, max iteration $T$}
    \BlankLine
    $\mathrm{\mathbf{I}}_\mathrm{Clean}^{(0)}\leftarrow \mathrm{\mathbf{I}}_\mathrm{Noisy}$
    
    \For{$t \gets 1$ to $T$} {
        $\mathrm{\mathbf{y}}^{(t)} \leftarrow \mathrm{\mathbf{I}}_\mathrm{Clean}^{(t-1)}$

        \eIf{$t=1$}
        {
            $\mathrm{\mathbf{I}}^{(t)}_{\mathrm{Res}} \leftarrow 0$
        }{
            $\mathrm{\mathbf{I}}^{(t)}_{\mathrm{Res}} \leftarrow \mathrm{\mathbf{\hat{I}}}^{(t-1)}_\mathrm{Clean} - \mathrm{\mathbf{\hat{I}}}^{(t-2)}_\mathrm{Clean}$
        }

        $\text{dilation} \leftarrow (t \mod 2 = 1)$
        
        $ \mathbf{I}_\mathrm{Clean}^{(t)} \leftarrow \text{DID-Block}(\mathbf{y}^{(t)}, \mathrm{\mathbf{I}}^{(t)}_{\mathrm{Res}}, \text{dilation})$
        
        \If{\text{criterion in~\eeqref{eq:stop_threshold} is met}}{
        $T \leftarrow t$
         
        }  
    }
    \Return Denoised image $\mathbf{\hat{I}}_\mathrm{Clean}^{(T)}$
    \caption{Dynamic Iteration Control (DIC)}
    \label{alg:DIC}
\end{algorithm}

\subsection{DIC Algorithm}
The overall inference algorithm of \framework{} is outlined in~\algref{alg:DIC}. The noisy input image is iteratively denoised for up to $T$ iterations. At each iteration, the dilation rate within the DID block is alternated to balance global and local context. Specifically, for odd-numbered iterations, the dilation rate is set to two to capture broader contextual regions. In contrast, during even-numbered iterations, the rate is reduced to one, enabling a focus on local details. If DIC is enabled and the stopping criterion is satisfied (see~\eeqref{eq:stop_threshold}), the iterative denoising process is terminated early.

\subsection{Additional Analysis on DIC}
To accelerate inference efficiency, 
we propose the Dynamic Iteration Control (DIC) mechanism, which adaptively determines the number of denoising iterations based on image content and noise characteristics. 
To comprehensively evaluate DIC, we compare two variants in the following subsections. 
One variant, referred to as Image-DIC, utilizes the residual of the denoised images, 
while the other variant, termed Kernel-DIC, utilizes the residual of the predicted kernels, as described in~\eeqref{eq:stop_threshold}.

\begin{table}[t]
\centering
    \caption{Comparison results on denoising performance depending on DIC strategies with equivalent average iterations. The results of Kernel-DIC are highlighted in \textbf{bold}.}
    \resizebox{0.475\textwidth}{!}{
\begin{tabular}{cccc}
\toprule[0.5pt]
\multirow{2}{*}{Methods} & Gaussian & Spatial Gaussian         & SIDD        \\ \cmidrule(lr){2-4} 
                         & PSNR$\uparrow$/SSIM$\uparrow$        & PSNR$\uparrow$/SSIM$\uparrow$    & PSNR$\uparrow$/SSIM$\uparrow$    \\ \midrule[0.2pt]
No-DIC                    & 31.89/0.9029     & 27.09/0.7949 & 29.42/0.6384 \\
Image-DIC                 & 31.75/0.9008     & 27.14/0.7976 & 29.85/0.6603 \\
\textbf{Kernel-DIC}       & \textbf{31.79/0.9010}     & \textbf{27.22/0.8025} & \textbf{30.09/0.6739} \\ \bottomrule[0.5pt]
\end{tabular}
    \label{tab:abl_kernel_fixed_iteration}
    }
\end{table}

\noindent\textbf{Comparison of Results with Equivalent Average Iterations.}
\ttabref{tab:abl_kernel_fixed_iteration} compares the denoising performance of three strategies: No-DIC, Image-DIC, and Kernel-DIC, while maintaining an equivalent average number of iterations by manually adjusting the threshold $\kappa$ in~\eeqref{eq:stop_threshold} for each test dataset. On the Gaussian noise dataset with $\sigma=15$, which represents an in-distribution scenario, No-DIC achieves the highest PSNR and SSIM. However, the differences compared to the DIC-based variants are marginal.

For the more challenging spatial Gaussian noise with $\sigma=55$, both DIC strategies lead to performance improvements. Image-DIC provides moderate gains, while Kernel-DIC offers further enhancement and outperforms No-DIC. It is worth noting that Urban100 is used for both synthetic noise settings. A similar trend is observed on the real-world SIDD dataset, where the use of DIC results in notable performance gains, with Kernel-DIC achieving the best results among the three.

These findings suggest that adaptive iteration strategies perform comparably to fixed-iteration methods under simple noise conditions. In contrast, under complex noise distributions, they provide clear benefits, with the kernel-residual-based DIC demonstrating the most consistent and robust improvements.

\noindent\textbf{Comparison of Image and Kernel-based DIC Results with Different Levels of Noise.}
In~\ffigref{fig:comparsion_with_image_DIC}, both Image-DIC and Kernel-DIC methods are evaluated under spatial Gaussian noise at varying magnitudes (\eg, $\sigma=\{45,50,55\}$) to determine which residual feature, image or kernel, better reflects noise characteristics. As the noise level increases from 45 to 55, the average number of iterations required by the kernel-DIC (Kernel-Residual) increases linearly. In contrast, the iterations for the image-DIC (Image-Residual) remain largely unchanged across noise levels. These findings indicate that kernel-DIC is more responsive to noise magnitude, allowing for adaptive inference and improved computational efficiency in noise-dependent scenarios.

\begin{figure}
\begin{center}
\centerline{\includegraphics[width=1\columnwidth]{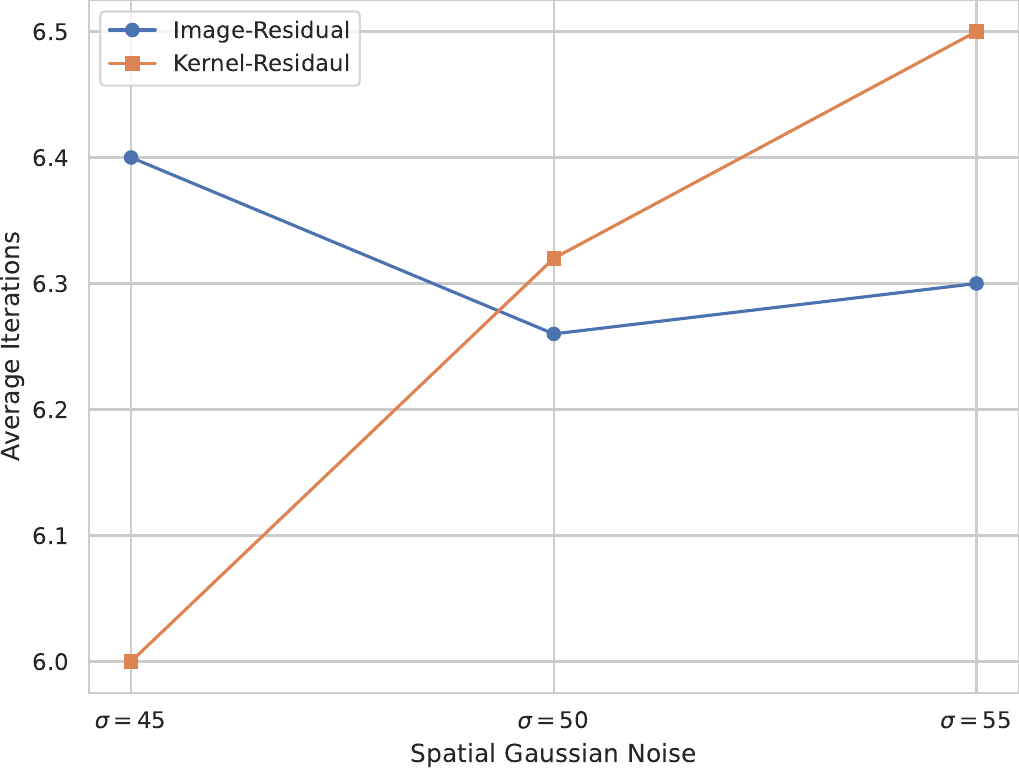}}
\caption{Comparison of results on averaged adaptive iterations with different levels of noise using image residual–based DIC (Image-DIC) and the kernel residual–based DIC (Kernel-DIC) approaches.}
\label{fig:comparsion_with_image_DIC}
\vspace{-10mm}
\end{center}
\end{figure}

\subsection{Visualization of Predicted Kernels}
\label{append_kernel_vis}
To better understand how our model adapts to different noise characteristics, in~\ffigref{fig:kernel_map}, we visualize the predicted kernel maps under a variety of noise types, including both synthetic and real-world degradations. For each example, we select representative pixel locations (highlighted with red dots) and display their corresponding denoising kernels. The predicted kernels vary in shape depending on both the spatial structure and the underlying noise distribution. Notably, in flat or homogeneous regions, the kernels exhibit near-uniform weights, enabling effective averaging to suppress noise. In contrast, in textured or edge regions, the kernels become more anisotropic, preserving local structures. Importantly, all predicted kernels are normalized to sum to one, effectively functioning as content-adaptive averaging filters. This regularization improves stability and prevents over-amplification of noise, especially under OOD settings. These visualizations highlight how our model generalizes across noise domains by dynamically modulating its receptive behavior based on local context.

\subsection{Additional Analysis on Iterative Method}
\noindent\textbf{Visualization of the Iterative Refinement.}
To effectively demonstrate how denoising kernels and denoised images evolve over iterations, we present results from all ten iterations ($T=10$). Specifically, the denoised images are shown in the upper row, while the corresponding averaged denoising kernels are displayed in the lower row for clarity. Gaussian and spatial Gaussian noise at levels $\sigma=50$ and $\sigma=55$, respectively, are applied to the CBSD68~\cite{cbsd} and Urban100~\cite{urban} datasets for synthetic noise removal. Additionally, the SIDD~\cite{sidd} and SIDD+~\cite{siddplus} datasets are used for evaluating real-world noise removal.

As illustrated in~\ffigref{fig:iterative_results_cbsd68} and~\ffigref{fig:iterative_results_urban100}, even when the type and level of noise degradation remain the same, the denoising kernels differ between datasets. This variation reflects the model’s ability to adapt to specific image characteristics such as textures, and highlights its capacity for input-dependent filter generation. Moreover, the model exhibits sensitivity to different noise types, generating distinct kernels based on noise characteristics, even when the image content is unchanged.

Similarly, under real-world noise conditions (see~\ffigref{fig:iterative_results_real_world}), the proposed framework dynamically adjusts to the unique properties of the input signal, demonstrating strong generalization capabilities in OOD scenarios. Notably, regardless of signal content or noise characteristics, the denoising kernels show progressive convergence across iterations, which reflects the stability and robustness of the iterative denoising process.

These observations confirm that the model is highly robust in OOD denoising tasks, consistently producing kernels that flexibly adapt to diverse image content, noise types, and intensity levels.

\subsection{Additional Denoising Results}
\noindent\textbf{Qualitative Comparison.}
We provide additional visual comparisons with other benchmark models for both synthetic and real-world noise removal in~\ffigref{fig:abl_synthetic_denoising_extension} and~\ffigref{fig:abl_real_world_denoising_extension}, respectively. For reference, the original clean images and their corresponding cropped regions of interest (ROI) are presented in~\ffigref{fig:abl_whole_gt_image}.

\noindent\textbf{Quantitative Comparison.}
We further evaluate the generalization capability of our method by comparing its denoising performance with several benchmark models, including DnCNN~\cite{dncnn}, SwinIR~\cite{swinir}, Restormer~\cite{restormer}, CODE~\cite{code}, and MaskedDenoising~\cite{masked_denoising}, across various synthetic noise types and levels. Results from CLIPDenoising~\cite{clip_denoising} are also included for reference. A detailed comparison is provided in~\ttabref{tab:abl_synthetic_denoising_extension}.

%%%%%%%%%%%%%%%%%%%%%%%%%%%%%%%%%%%%%%%%%%%%%%%%%%%%%%%%%%%%%%%%%%%%%%%%%%%%%%%%%%%%%%%%%%%%%%%%%%%%%%%%%%%%%%%%%%%%%%%%%%%%%%%%%%%%%%%%%%%%%%%

\begin{table*}[t]
  \centering
  \caption{Quantitative results of denoising performance on CBSD68, McMaster, Kodak24 and Urban100 with regard to varied synthetic OOD noises in terms of PSNR$\uparrow$ and SSIM$\uparrow$. All methods are trained with Gaussian noise with a level of $\sigma=15$. The symbol $\dagger$ denotes that our model utilizes the proposed DIC during inference. The best and second-best results are highlighted in \textbf{bold} and \underline{underline}.}
  \resizebox{1.0\textwidth}{!}{
    \begin{tabular}{cc|cc|cccccccccccccc}
    \toprule
    \multicolumn{1}{r}{\multirow{2}{*}{Noise Types}} & \multirow{2}{*}{Datasets} & \multicolumn{2}{c|}{ClipDenoising} & \multicolumn{2}{c}{DnCNN} & \multicolumn{2}{c}{SwinIR} & \multicolumn{2}{c}{Restormer} & \multicolumn{2}{c}{CODE} & \multicolumn{2}{c}{MaskedDenoising} & \multicolumn{2}{c}{\textbf{\textbf{Ours}}$^\dagger$} & \multicolumn{2}{c}{\textbf{Ours}} \\
\cmidrule(lr){3-4} \cmidrule(lr){5-6} \cmidrule(lr){7-8} \cmidrule(lr){9-10} \cmidrule(lr){11-12} \cmidrule(lr){13-14} \cmidrule(lr){15-16} \cmidrule(lr){17-18}           &       & \multicolumn{2}{c|}{PSNR$\uparrow$/SSIM$\uparrow$}  & \multicolumn{2}{c}{PSNR$\uparrow$/SSIM$\uparrow$}  & \multicolumn{2}{c}{PSNR$\uparrow$/SSIM$\uparrow$}  & \multicolumn{2}{c}{PSNR$\uparrow$/SSIM$\uparrow$}  & \multicolumn{2}{c}{PSNR$\uparrow$/SSIM$\uparrow$}  & \multicolumn{2}{c}{PSNR$\uparrow$/SSIM$\uparrow$}  & \multicolumn{2}{c}{PSNR$\uparrow$/SSIM$\uparrow$}  & \multicolumn{2}{c}{PSNR$\uparrow$/SSIM$\uparrow$} \\
    \midrule
    \multicolumn{1}{r}{\multirow{4}{*}{\makecell{Gaussian \\ $\sigma=15$}}}  &  CBSD68  & \multicolumn{2}{c|}{33.61/0.9273} & \multicolumn{2}{c}{33.64/0.9271} & \multicolumn{2}{c}{\underline{34.00}/0.9319} & \multicolumn{2}{c}{33.99/\underline{0.9319}} & \multicolumn{2}{c}{\textbf{34.10}/\textbf{0.9339}} & \multicolumn{2}{c}{30.78/0.8891} & \multicolumn{2}{c}{32.24/0.8939} & \multicolumn{2}{c}{32.18/0.8915} \\
           &  MCMaster  & \multicolumn{2}{c|}{33.49/0.8932} & \multicolumn{2}{c}{34.30/0.9036} & \multicolumn{2}{c}{\underline{34.95}/\underline{0.9117}} & \multicolumn{2}{c}{34.83/0.9093} & \multicolumn{2}{c}{\textbf{35.11}/\textbf{0.9276}} & \multicolumn{2}{c}{30.90/0.8502} & \multicolumn{2}{c}{32.52/0.8776} & \multicolumn{2}{c}{32.11/0.8680} \\
           &  Kodak24  & \multicolumn{2}{c|}{34.27/0.9193} & \multicolumn{2}{c}{34.42/0.9209} & \multicolumn{2}{c}{\textbf{34.99}/\underline{0.9285}} & \multicolumn{2}{c}{\underline{34.98}/\textbf{0.9287}} & \multicolumn{2}{c}{34.95/0.9285} & \multicolumn{2}{c}{31.41/0.8833} & \multicolumn{2}{c}{32.93/0.8867} & \multicolumn{2}{c}{32.87/0.8839} \\
           &  Urban100  & \multicolumn{2}{c|}{32.77/0.9179} & \multicolumn{2}{c}{33.64/0.9264} & \multicolumn{2}{c}{\underline{34.49}/\underline{0.9338}} & \multicolumn{2}{c}{34.45/0.9338} & \multicolumn{2}{c}{\textbf{34.59}/\textbf{0.9492}} & \multicolumn{2}{c}{29.32/0.8995} & \multicolumn{2}{c}{31.52/0.8947} & \multicolumn{2}{c}{31.42/0.8929} \\
    \midrule
    \midrule
    \multicolumn{1}{r}{\multirow{4}{*}{\makecell{Gaussian \\ $\sigma=25$}}}  &  CBSD68  & \multicolumn{2}{c|}{30.51/0.8718} & \multicolumn{2}{c}{24.89/0.6010} & \multicolumn{2}{c}{24.29/0.5639} & \multicolumn{2}{c}{27.51/0.7151} & \multicolumn{2}{c}{24.58/0.5631} & \multicolumn{2}{c}{28.20/0.8202} & \multicolumn{2}{c}{\underline{29.64}/\underline{0.8376}} & \multicolumn{2}{c}{\textbf{29.80}/\textbf{0.8435}} \\
           &  MCMaster  & \multicolumn{2}{c|}{30.62/0.8319} & \multicolumn{2}{c}{25.63/0.5818} & \multicolumn{2}{c}{25.03/0.5468} & \multicolumn{2}{c}{28.07/0.6910} & \multicolumn{2}{c}{25.21/0.5395} & \multicolumn{2}{c}{28.99/0.7971} & \multicolumn{2}{c}{\textbf{30.16}/\textbf{0.8218}} & \multicolumn{2}{c}{\underline{30.10}/\underline{0.8195}} \\
           &  Kodak24  & \multicolumn{2}{c|}{31.40/0.8681} & \multicolumn{2}{c}{24.83/0.5388} & \multicolumn{2}{c}{24.26/0.5023} & \multicolumn{2}{c}{27.87/0.6731} & \multicolumn{2}{c}{24.57/0.5067} & \multicolumn{2}{c}{28.85/0.8004} & \multicolumn{2}{c}{\underline{30.36}/\underline{0.8296}} & \multicolumn{2}{c}{\textbf{30.64}/\textbf{0.8410}} \\
           &  Urban100  & \multicolumn{2}{c|}{30.05/0.8761} & \multicolumn{2}{c}{25.28/0.6476} & \multicolumn{2}{c}{24.66/0.6147} & \multicolumn{2}{c}{28.06/0.7530} & \multicolumn{2}{c}{24.99/0.6195} & \multicolumn{2}{c}{27.51/0.8419} & \multicolumn{2}{c}{\underline{29.17}/\underline{0.8534}} & \multicolumn{2}{c}{\textbf{29.28}/\textbf{0.8571}} \\
    \midrule
    \multicolumn{1}{r}{\multirow{4}{*}{\makecell{Spatial \\ Gaussian \\ $\sigma=45$}}}  &  CBSD68  & \multicolumn{2}{c|}{29.34/0.8488} & \multicolumn{2}{c}{28.19/0.7907} & \multicolumn{2}{c}{27.27/0.7391} & \multicolumn{2}{c}{24.14/0.6686} & \multicolumn{2}{c}{27.27/0.7353} & \multicolumn{2}{c}{28.13/0.8181} & \multicolumn{2}{c}{\underline{29.01}/\underline{0.8368}} & \multicolumn{2}{c}{\textbf{29.20}/\textbf{0.8440}} \\
           &  MCMaster  & \multicolumn{2}{c|}{29.79/0.8236} & \multicolumn{2}{c}{\underline{28.68}/0.7707} & \multicolumn{2}{c}{27.79/0.7189} & \multicolumn{2}{c}{23.93/0.6059} & \multicolumn{2}{c}{27.55/0.6890} & \multicolumn{2}{c}{28.43/0.7778} & \multicolumn{2}{c}{\textbf{29.21}/\textbf{0.8035}} & \multicolumn{2}{c}{28.63/\underline{0.7860}} \\
           &  Kodak24  & \multicolumn{2}{c|}{29.97/0.8377} & \multicolumn{2}{c}{28.32/0.7591} & \multicolumn{2}{c}{27.34/0.7006} & \multicolumn{2}{c}{22.98/0.6192} & \multicolumn{2}{c}{27.41/0.7040} & \multicolumn{2}{c}{28.73/0.8105} & \multicolumn{2}{c}{\underline{29.32}/\underline{0.8140}} & \multicolumn{2}{c}{\textbf{29.79}/\textbf{0.8315}} \\
           &  Urban100  & \multicolumn{2}{c|}{29.38/0.8633} & \multicolumn{2}{c}{28.61/0.8148} & \multicolumn{2}{c}{27.64/0.7681} & \multicolumn{2}{c}{25.55/0.7013} & \multicolumn{2}{c}{27.54/0.7715} & \multicolumn{2}{c}{27.33/0.8425} & \multicolumn{2}{c}{\underline{28.77}/\underline{0.8511}} & \multicolumn{2}{c}{\textbf{28.98}/\textbf{0.8619}} \\
    \midrule
    \multicolumn{1}{r}{\multirow{4}{*}{\makecell{Spatial \\ Gaussian \\ $\sigma=50$}}}  &  CBSD68  & \multicolumn{2}{c|}{28.44/0.8263} & \multicolumn{2}{c}{26.98/0.7446} & \multicolumn{2}{c}{26.13/0.6918} & \multicolumn{2}{c}{23.72/0.6320} & \multicolumn{2}{c}{26.15/0.6894} & \multicolumn{2}{c}{27.43/0.7954} & \multicolumn{2}{c}{\underline{28.31}/\underline{0.8164}} & \multicolumn{2}{c}{\textbf{28.51}/\textbf{0.8250}} \\
           &  MCMaster  & \multicolumn{2}{c|}{29.12/0.8047} & \multicolumn{2}{c}{27.52/0.7231} & \multicolumn{2}{c}{26.64/0.6678} & \multicolumn{2}{c}{23.49/0.5728} & \multicolumn{2}{c}{26.48/0.6439} & \multicolumn{2}{c}{27.82/0.7571} & \multicolumn{2}{c}{\textbf{29.57}/\textbf{0.8256}} & \multicolumn{2}{c}{\underline{29.01}/\underline{0.8095}} \\
           &  Kodak24  & \multicolumn{2}{c|}{29.08/0.8149} & \multicolumn{2}{c}{27.06/0.7063} & \multicolumn{2}{c}{26.17/0.6480} & \multicolumn{2}{c}{22.84/0.5807} & \multicolumn{2}{c}{26.26/0.6520} & \multicolumn{2}{c}{28.00/0.7854} & \multicolumn{2}{c}{\underline{28.65}/\underline{0.7915}} & \multicolumn{2}{c}{\textbf{29.13}/\textbf{0.8128}} \\
           &  Urban100  & \multicolumn{2}{c|}{28.56/0.8438} & \multicolumn{2}{c}{27.38/0.7740} & \multicolumn{2}{c}{26.47/0.7268} & \multicolumn{2}{c}{24.78/0.6688} & \multicolumn{2}{c}{26.41/0.7302} & \multicolumn{2}{c}{26.77/0.8224} & \multicolumn{2}{c}{\underline{28.12}/\underline{0.8364}} & \multicolumn{2}{c}{\textbf{28.37}/\textbf{0.8483}} \\
    \midrule
    \multicolumn{1}{r}{\multirow{4}{*}{\makecell{Poisson \\ $\alpha=2.5$}}}  &  CBSD68  & \multicolumn{2}{c|}{29.89/0.8731} & \multicolumn{2}{c}{24.03/0.6261} & \multicolumn{2}{c}{23.67/0.6045} & \multicolumn{2}{c}{25.67/0.6941} & \multicolumn{2}{c}{23.99/0.6064} & \multicolumn{2}{c}{27.69/0.8024} & \multicolumn{2}{c}{\underline{29.21}/\underline{0.8425}} & \multicolumn{2}{c}{\textbf{29.36}/\textbf{0.8477}} \\
           &  MCMaster  & \multicolumn{2}{c|}{30.88/0.8628} & \multicolumn{2}{c}{24.94/0.6627} & \multicolumn{2}{c}{24.45/0.6460} & \multicolumn{2}{c}{25.78/0.6939} & \multicolumn{2}{c}{24.81/0.5906} & \multicolumn{2}{c}{28.42/0.7224} & \multicolumn{2}{c}{\underline{30.33}/\underline{0.8542}} & \multicolumn{2}{c}{\textbf{30.38}/\textbf{0.8565}} \\
           &  Kodak24  & \multicolumn{2}{c|}{30.77/0.8655} & \multicolumn{2}{c}{23.94/0.5605} & \multicolumn{2}{c}{23.58/0.5406} & \multicolumn{2}{c}{25.96/0.6440} & \multicolumn{2}{c}{23.94/0.5471} & \multicolumn{2}{c}{28.28/0.7796} & \multicolumn{2}{c}{\underline{30.08}/\underline{0.8359}} & \multicolumn{2}{c}{\textbf{30.30}/\textbf{0.8440}} \\
           &  Urban100  & \multicolumn{2}{c|}{29.44/0.8840} & \multicolumn{2}{c}{23.61/0.6537} & \multicolumn{2}{c}{23.24/0.6390} & \multicolumn{2}{c}{25.30/0.7044} & \multicolumn{2}{c}{23.65/0.6395} & \multicolumn{2}{c}{26.85/0.8125} & \multicolumn{2}{c}{\underline{28.87}/\underline{0.8724}} & \multicolumn{2}{c}{\textbf{28.94}/\textbf{0.8762}} \\
    \midrule
    \multicolumn{1}{r}{\multirow{4}{*}{\makecell{Poisson \\ $\alpha=3.0$}}}  &  CBSD68  & \multicolumn{2}{c|}{28.68/0.8457} & \multicolumn{2}{c}{21.36/0.5149} & \multicolumn{2}{c}{21.27/0.4988} & \multicolumn{2}{c}{23.53/0.6172} & \multicolumn{2}{c}{21.65/0.5023} & \multicolumn{2}{c}{25.79/0.7141} & \multicolumn{2}{c}{\underline{28.15}/\underline{0.8153}} & \multicolumn{2}{c}{\textbf{28.37}/\textbf{0.8243}} \\
           &  MCMaster  & \multicolumn{2}{c|}{29.80/0.8429} & \multicolumn{2}{c}{22.27/0.5816} & \multicolumn{2}{c}{22.11/0.5725} & \multicolumn{2}{c}{23.59/0.6322} & \multicolumn{2}{c}{22.48/0.5163} & \multicolumn{2}{c}{26.59/0.6416} & \multicolumn{2}{c}{\underline{29.44}/\underline{0.8355}} & \multicolumn{2}{c}{\textbf{29.54}/\textbf{0.8398}} \\
           &  Kodak24  & \multicolumn{2}{c|}{29.56/0.8382} & \multicolumn{2}{c}{21.16/0.4448} & \multicolumn{2}{c}{21.09/0.4321} & \multicolumn{2}{c}{23.88/0.5636} & \multicolumn{2}{c}{21.49/0.4376} & \multicolumn{2}{c}{26.04/0.6685} & \multicolumn{2}{c}{\underline{29.03}/\underline{0.8068}} & \multicolumn{2}{c}{\textbf{29.30}/\textbf{0.8213}} \\
           &  Urban100  & \multicolumn{2}{c|}{28.22/0.8614} & \multicolumn{2}{c}{21.02/0.5664} & \multicolumn{2}{c}{20.94/0.5574} & \multicolumn{2}{c}{22.87/0.6267} & \multicolumn{2}{c}{21.39/0.5590} & \multicolumn{2}{c}{25.25/0.7341} & \multicolumn{2}{c}{\underline{27.80}/\underline{0.8517}} & \multicolumn{2}{c}{\textbf{27.94}/\textbf{0.8596}} \\
    \midrule
    \multicolumn{1}{r}{\multirow{4}{*}{\makecell{Salt \\ \& \\ Pepper \\ $d=0.012$}}}  &  CBSD68  & \multicolumn{2}{c|}{31.96/0.8900} & \multicolumn{2}{c}{26.63/0.7968} & \multicolumn{2}{c}{25.51/0.7654} & \multicolumn{2}{c}{25.89/0.7788} & \multicolumn{2}{c}{26.50/0.7727} & \multicolumn{2}{c}{30.49/0.8623} & \multicolumn{2}{c}{\textbf{34.94}/\textbf{0.9355}} & \multicolumn{2}{c}{\underline{34.39}/\underline{0.9256}} \\
           &  MCMaster  & \multicolumn{2}{c|}{31.90/0.8633} & \multicolumn{2}{c}{25.51/0.7606} & \multicolumn{2}{c}{25.00/0.7420} & \multicolumn{2}{c}{25.33/0.7462} & \multicolumn{2}{c}{25.75/0.7149} & \multicolumn{2}{c}{30.10/0.7976} & \multicolumn{2}{c}{\textbf{33.81}/\textbf{0.9059}} & \multicolumn{2}{c}{\underline{33.13}/\underline{0.8948}} \\
           &  Kodak24  & \multicolumn{2}{c|}{32.62/0.8805} & \multicolumn{2}{c}{26.97/0.7764} & \multicolumn{2}{c}{25.75/0.7402} & \multicolumn{2}{c}{26.18/0.7547} & \multicolumn{2}{c}{26.92/0.7536} & \multicolumn{2}{c}{31.16/0.8619} & \multicolumn{2}{c}{\textbf{35.54}/\textbf{0.9244}} & \multicolumn{2}{c}{\underline{35.13}/\underline{0.9168}} \\
           &  Urban100  & \multicolumn{2}{c|}{31.50/0.9009} & \multicolumn{2}{c}{26.05/0.8146} & \multicolumn{2}{c}{25.15/0.7923} & \multicolumn{2}{c}{25.62/0.7993} & \multicolumn{2}{c}{26.48/0.8030} & \multicolumn{2}{c}{29.08/0.8802} & \multicolumn{2}{c}{\textbf{33.36}/\textbf{0.9290}} & \multicolumn{2}{c}{\underline{32.94}/\underline{0.9229}} \\
    \midrule
    \multicolumn{1}{r}{\multirow{4}{*}{\makecell{Salt \\ \& \\ Pepper \\ $d=0.016$}}}  &  CBSD68  & \multicolumn{2}{c|}{30.85/0.8700} & \multicolumn{2}{c}{25.18/0.7518} & \multicolumn{2}{c}{24.23/0.7174} & \multicolumn{2}{c}{24.57/0.7256} & \multicolumn{2}{c}{25.13/0.7269} & \multicolumn{2}{c}{30.13/0.8537} & \multicolumn{2}{c}{\textbf{34.29}/\textbf{0.9285}} & \multicolumn{2}{c}{\underline{33.84}/\underline{0.9191}} \\
           &  MCMaster  & \multicolumn{2}{c|}{30.83/0.8377} & \multicolumn{2}{c}{24.09/0.7094} & \multicolumn{2}{c}{23.69/0.6878} & \multicolumn{2}{c}{24.01/0.6883} & \multicolumn{2}{c}{24.41/0.6679} & \multicolumn{2}{c}{29.68/0.7856} & \multicolumn{2}{c}{\textbf{33.05}/\textbf{0.8953}} & \multicolumn{2}{c}{\underline{32.79}/\underline{0.8897}} \\
           &  Kodak24  & \multicolumn{2}{c|}{31.48/0.8593} & \multicolumn{2}{c}{25.43/0.7249} & \multicolumn{2}{c}{24.42/0.6850} & \multicolumn{2}{c}{24.78/0.6932} & \multicolumn{2}{c}{25.47/0.7010} & \multicolumn{2}{c}{30.82/0.8532} & \multicolumn{2}{c}{\textbf{34.91}/\textbf{0.9171}} & \multicolumn{2}{c}{\underline{34.58}/\underline{0.9102}} \\
           &  Urban100  & \multicolumn{2}{c|}{30.57/0.8846} & \multicolumn{2}{c}{24.66/0.7730} & \multicolumn{2}{c}{23.89/0.7479} & \multicolumn{2}{c}{24.36/0.7537} & \multicolumn{2}{c}{25.13/0.7618} & \multicolumn{2}{c}{28.76/0.8713} & \multicolumn{2}{c}{\textbf{32.87}/\textbf{0.9252}} & \multicolumn{2}{c}{\underline{32.53}/\underline{0.9191}} \\
    \midrule
    \multicolumn{1}{r}{\multirow{4}{*}{\makecell{Speckle \\ ${\sigma}^2=0.02$}}}  &  CBSD68  & \multicolumn{2}{c|}{31.81/0.9038} & \multicolumn{2}{c}{29.72/0.8308} & \multicolumn{2}{c}{28.88/0.8100} & \multicolumn{2}{c}{29.15/0.8277} & \multicolumn{2}{c}{29.32/0.8143} & \multicolumn{2}{c}{29.91/0.8752} & \multicolumn{2}{c}{\textbf{31.17}/\textbf{0.8910}} & \multicolumn{2}{c}{\underline{31.16}/\underline{0.8897}} \\
           &  MCMaster  & \multicolumn{2}{c|}{32.28/0.8703} & \multicolumn{2}{c}{30.32/0.8156} & \multicolumn{2}{c}{29.17/0.7946} & \multicolumn{2}{c}{28.89/0.8003} & \multicolumn{2}{c}{29.87/0.7573} & \multicolumn{2}{c}{30.47/0.8090} & \multicolumn{2}{c}{\textbf{31.77}/\textbf{0.8798}} & \multicolumn{2}{c}{\underline{31.73}/\underline{0.8785}} \\
           &  Kodak24  & \multicolumn{2}{c|}{32.69/0.9048} & \multicolumn{2}{c}{30.34/0.8173} & \multicolumn{2}{c}{29.39/0.7907} & \multicolumn{2}{c}{29.73/0.8129} & \multicolumn{2}{c}{30.04/0.8035} & \multicolumn{2}{c}{30.65/0.8739} & \multicolumn{2}{c}{\underline{31.94}/\textbf{0.8836}} & \multicolumn{2}{c}{\textbf{31.96}/\underline{0.8828}} \\
           &  Urban100  & \multicolumn{2}{c|}{30.94/0.9043} & \multicolumn{2}{c}{28.42/0.8130} & \multicolumn{2}{c}{27.50/0.7930} & \multicolumn{2}{c}{28.22/0.8100} & \multicolumn{2}{c}{28.03/0.7959} & \multicolumn{2}{c}{28.60/0.8832} & \multicolumn{2}{c}{\textbf{30.39}/\textbf{0.8981}} & \multicolumn{2}{c}{\underline{30.31}/\underline{0.8977}} \\
    \midrule
    \multicolumn{1}{r}{\multirow{4}{*}{\makecell{Speckle \\ ${\sigma}^2=0.03$}}}  &  CBSD68  & \multicolumn{2}{c|}{30.49/0.8863} & \multicolumn{2}{c}{26.68/0.7546} & \multicolumn{2}{c}{25.98/0.7363} & \multicolumn{2}{c}{26.84/0.7668} & \multicolumn{2}{c}{26.47/0.7440} & \multicolumn{2}{c}{29.00/0.8509} & \multicolumn{2}{c}{\underline{29.97}/\underline{0.8697}} & \multicolumn{2}{c}{\textbf{30.03}/\textbf{0.8712}} \\
           &  MCMaster  & \multicolumn{2}{c|}{31.30/0.8578} & \multicolumn{2}{c}{27.19/0.7492} & \multicolumn{2}{c}{26.29/0.7325} & \multicolumn{2}{c}{26.82/0.7526} & \multicolumn{2}{c}{27.03/0.6946} & \multicolumn{2}{c}{29.69/0.7774} & \multicolumn{2}{c}{\underline{30.92}/\textbf{0.8672}} & \multicolumn{2}{c}{\textbf{30.93}/\underline{0.8667}} \\
           &  Kodak24  & \multicolumn{2}{c|}{31.40/0.8878} & \multicolumn{2}{c}{27.06/0.7232} & \multicolumn{2}{c}{26.21/0.6947} & \multicolumn{2}{c}{27.29/0.7382} & \multicolumn{2}{c}{26.86/0.7100} & \multicolumn{2}{c}{29.74/0.8479} & \multicolumn{2}{c}{\underline{30.77}/\underline{0.8619}} & \multicolumn{2}{c}{\textbf{30.89}/\textbf{0.8656}} \\
           &  Urban100  & \multicolumn{2}{c|}{29.69/0.8889} & \multicolumn{2}{c}{25.33/0.7398} & \multicolumn{2}{c}{24.68/0.7256} & \multicolumn{2}{c}{25.86/0.7529} & \multicolumn{2}{c}{25.24/0.7300} & \multicolumn{2}{c}{27.65/0.8466} & \multicolumn{2}{c}{\textbf{29.30}/\underline{0.8840}} & \multicolumn{2}{c}{\underline{29.28}/\textbf{0.8852}} \\
    \midrule
    \multicolumn{1}{r}{\multirow{4}{*}{\makecell{Mixture \\ Level 1}}}  &  CBSD68  & \multicolumn{2}{c|}{30.91/0.8930} & \multicolumn{2}{c}{27.43/0.7713} & \multicolumn{2}{c}{27.03/0.7476} & \multicolumn{2}{c}{28.44/0.8090} & \multicolumn{2}{c}{27.14/0.7372} & \multicolumn{2}{c}{29.08/\textbf{0.8710}} & \multicolumn{2}{c}{\underline{30.27}/0.8701} & \multicolumn{2}{c}{\textbf{30.32}/\underline{0.8707}} \\
           &  MCMaster  & \multicolumn{2}{c|}{31.62/0.8664} & \multicolumn{2}{c}{27.88/0.7494} & \multicolumn{2}{c}{27.53/0.7306} & \multicolumn{2}{c}{28.54/0.7707} & \multicolumn{2}{c}{27.61/0.7137} & \multicolumn{2}{c}{29.85/0.8104} & \multicolumn{2}{c}{\underline{31.05}/\underline{0.8606}} & \multicolumn{2}{c}{\textbf{31.06}/\textbf{0.8610}} \\
           &  Kodak24  & \multicolumn{2}{c|}{31.72/0.8874} & \multicolumn{2}{c}{27.66/0.7296} & \multicolumn{2}{c}{27.26/0.7053} & \multicolumn{2}{c}{29.03/0.7872} & \multicolumn{2}{c}{27.44/0.6976} & \multicolumn{2}{c}{29.91/\textbf{0.8663}} & \multicolumn{2}{c}{\underline{31.00}/0.8627} & \multicolumn{2}{c}{\textbf{31.11}/\underline{0.8658}} \\
           &  Urban100  & \multicolumn{2}{c|}{30.40/0.8928} & \multicolumn{2}{c}{27.13/0.7692} & \multicolumn{2}{c}{26.73/0.7482} & \multicolumn{2}{c}{28.37/0.8091} & \multicolumn{2}{c}{26.98/0.7539} & \multicolumn{2}{c}{27.97/\textbf{0.8799}} & \multicolumn{2}{c}{\underline{29.77}/0.8794} & \multicolumn{2}{c}{\textbf{29.77}/\underline{0.8798}} \\
    \midrule
    \multicolumn{1}{r}{\multirow{4}{*}{\makecell{Mixture \\ Level 2}}}  &  CBSD68  & \multicolumn{2}{c|}{30.31/0.8816} & \multicolumn{2}{c}{25.86/0.6960} & \multicolumn{2}{c}{25.46/0.6668} & \multicolumn{2}{c}{27.42/0.7623} & \multicolumn{2}{c}{25.62/0.6582} & \multicolumn{2}{c}{28.44/0.8545} & \multicolumn{2}{c}{\underline{29.68}/\underline{0.8576}} & \multicolumn{2}{c}{\textbf{29.76}/\textbf{0.8597}} \\
           &  MCMaster  & \multicolumn{2}{c|}{31.07/0.8537} & \multicolumn{2}{c}{26.43/0.6856} & \multicolumn{2}{c}{26.05/0.6658} & \multicolumn{2}{c}{27.48/0.7240} & \multicolumn{2}{c}{26.18/0.6583} & \multicolumn{2}{c}{29.36/0.8011} & \multicolumn{2}{c}{\underline{30.58}/\underline{0.8499}} & \multicolumn{2}{c}{\textbf{30.61}/\textbf{0.8505}} \\
           &  Kodak24  & \multicolumn{2}{c|}{31.15/0.8769} & \multicolumn{2}{c}{25.96/0.6414} & \multicolumn{2}{c}{25.57/0.6126} & \multicolumn{2}{c}{27.97/0.7321} & \multicolumn{2}{c}{25.77/0.6073} & \multicolumn{2}{c}{29.22/0.8446} & \multicolumn{2}{c}{\underline{30.39}/\underline{0.8486}} & \multicolumn{2}{c}{\textbf{30.58}/\textbf{0.8557}} \\
           &  Urban100  & \multicolumn{2}{c|}{29.81/0.8834} & \multicolumn{2}{c}{25.63/0.7060} & \multicolumn{2}{c}{25.21/0.6825} & \multicolumn{2}{c}{27.31/0.7668} & \multicolumn{2}{c}{25.48/0.6918} & \multicolumn{2}{c}{27.47/0.8645} & \multicolumn{2}{c}{\underline{29.17}/\underline{0.8691}} & \multicolumn{2}{c}{\textbf{29.23}/\textbf{0.8711}} \\
    \midrule
    \multicolumn{1}{r}{\multirow{4}{*}{\makecell{Mixture \\ Level 3}}}  &  CBSD68  & \multicolumn{2}{c|}{29.21/0.8569} & \multicolumn{2}{c}{23.20/0.5652} & \multicolumn{2}{c}{22.93/0.5375} & \multicolumn{2}{c}{25.11/0.6519} & \multicolumn{2}{c}{23.16/0.5332} & \multicolumn{2}{c}{26.95/0.7905} & \multicolumn{2}{c}{\underline{28.69}/\underline{0.8338}} & \multicolumn{2}{c}{\textbf{28.83}/\textbf{0.8398}} \\
           &  MCMaster  & \multicolumn{2}{c|}{30.01/0.8264} & \multicolumn{2}{c}{23.77/0.5665} & \multicolumn{2}{c}{23.56/0.5510} & \multicolumn{2}{c}{25.24/0.6294} & \multicolumn{2}{c}{23.72/0.5367} & \multicolumn{2}{c}{27.99/0.7488} & \multicolumn{2}{c}{\underline{29.71}/\underline{0.8269}} & \multicolumn{2}{c}{\textbf{29.78}/\textbf{0.8298}} \\
           &  Kodak24  & \multicolumn{2}{c|}{30.08/0.8537} & \multicolumn{2}{c}{23.18/0.5028} & \multicolumn{2}{c}{22.92/0.4750} & \multicolumn{2}{c}{25.49/0.6054} & \multicolumn{2}{c}{23.16/0.4745} & \multicolumn{2}{c}{27.50/0.7634} & \multicolumn{2}{c}{\underline{29.51}/\underline{0.8292}} & \multicolumn{2}{c}{\textbf{29.70}/\textbf{0.8375}} \\
           &  Urban100  & \multicolumn{2}{c|}{28.70/0.8632} & \multicolumn{2}{c}{23.03/0.5976} & \multicolumn{2}{c}{22.78/0.5766} & \multicolumn{2}{c}{24.98/0.6720} & \multicolumn{2}{c}{23.07/0.5857} & \multicolumn{2}{c}{26.29/0.8093} & \multicolumn{2}{c}{\underline{28.18}/\underline{0.8503}} & \multicolumn{2}{c}{\textbf{28.29}/\textbf{0.8550}} \\
    \bottomrule
    \end{tabular}%
    }
  \label{tab:abl_synthetic_denoising_extension}%
\end{table*}%

\begin{figure*}[ht]
\begin{center}
\centerline{\includegraphics[width=0.7\textwidth]{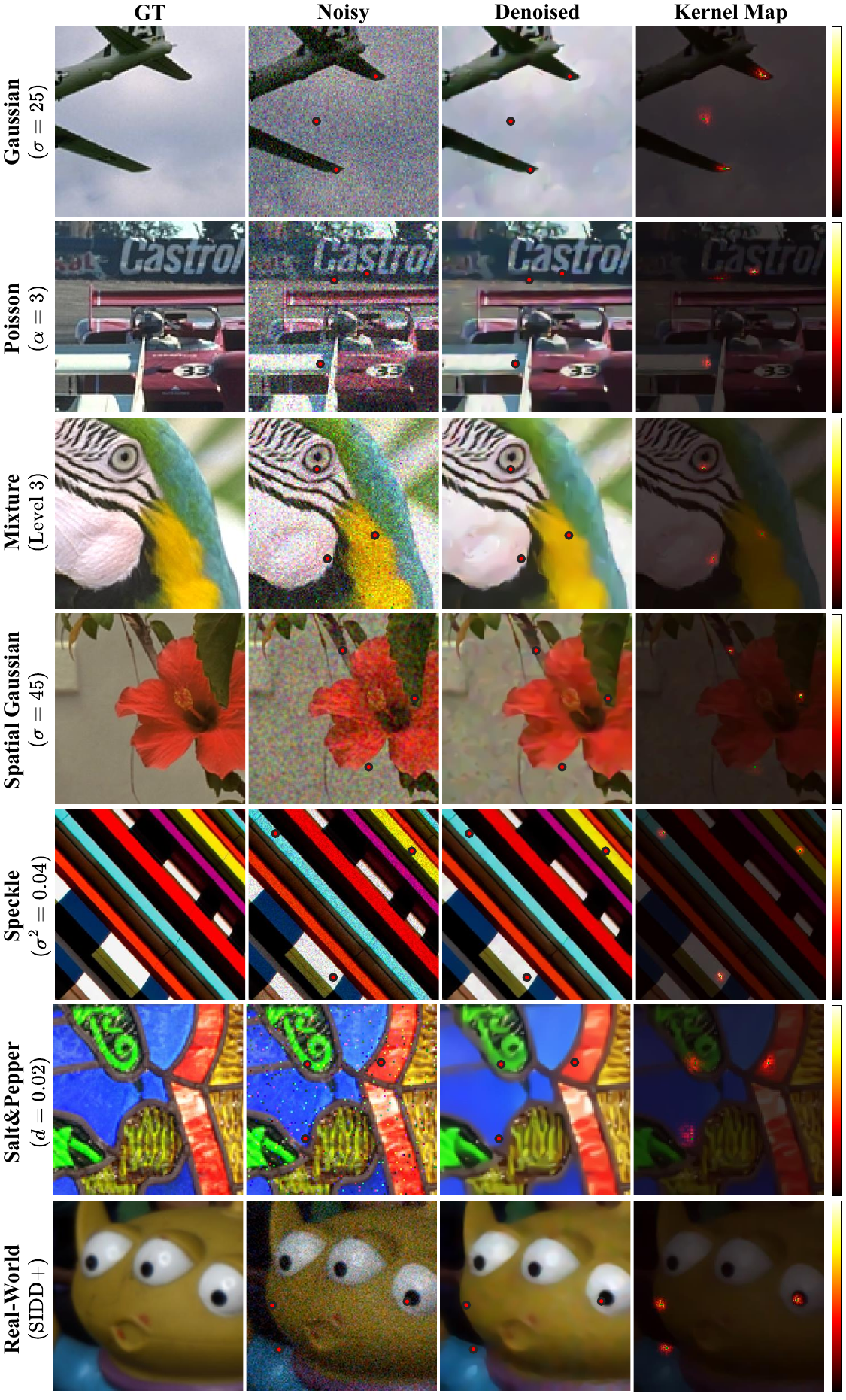}}
\caption{Visualizations of predicted kernel maps across diverse noise types.
For each noise condition, we display the ground-truth (GT), noisy input, denoised output, and the predicted kernel map. Red dots in the \textbf{Noisy} and \textbf{Denoised} columns indicate reference pixels used to visualize the corresponding denoising kernels. All kernel maps are normalized to sum to one, functioning as adaptive averaging filters.}
\label{fig:kernel_map}
\vspace{-10mm}
\end{center}
\end{figure*}

\begin{figure*}[ht]
\begin{center}
\centerline{\includegraphics[width=2.15\columnwidth]{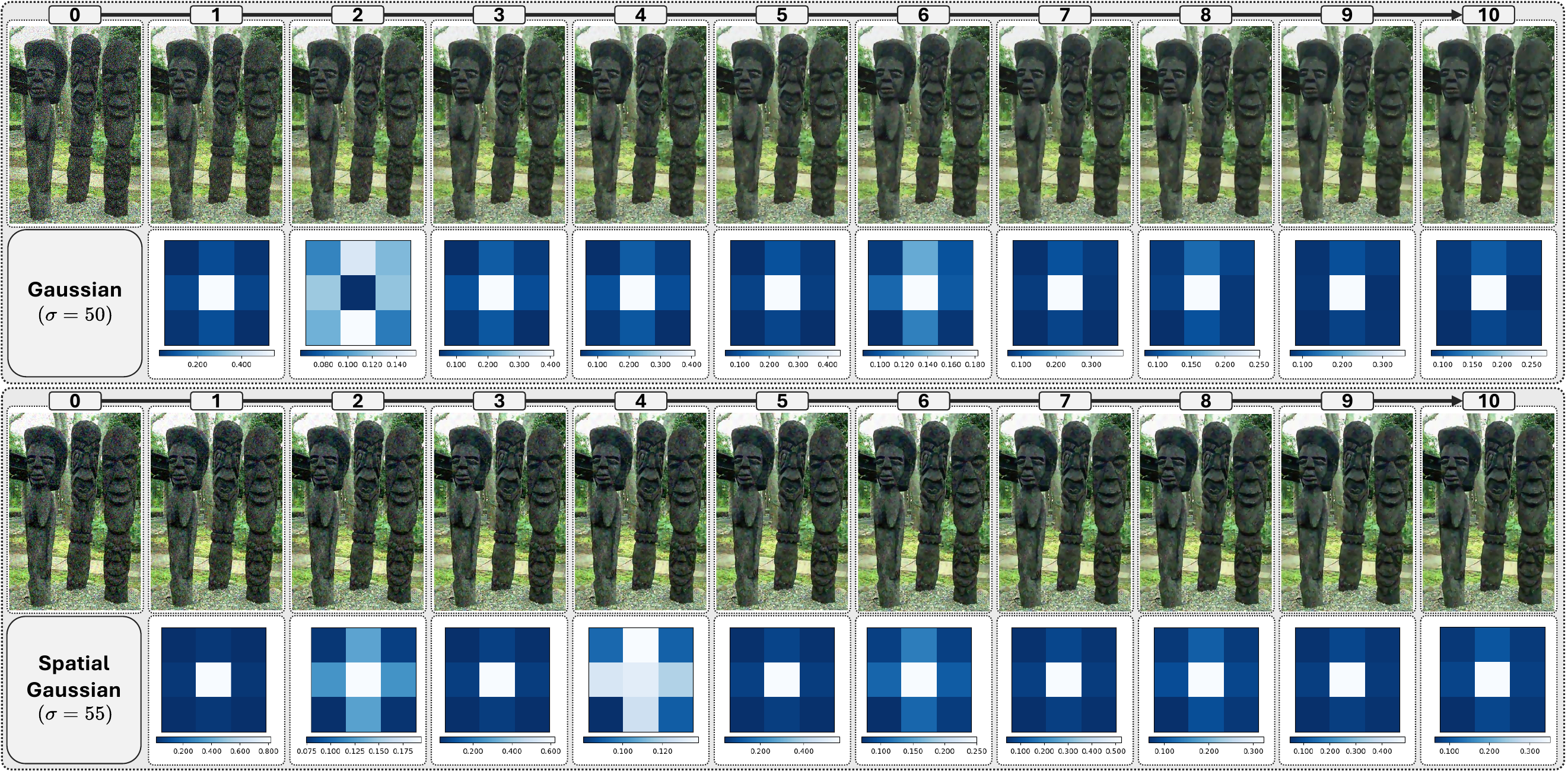}}
\caption{Comparison of results illustrating how denoising kernels and denoised images evolve through iterations, along with varying degradation types and levels on the CBSD68~\cite{cbsd} dataset. Please zoom in for a more detailed comparison.}
\label{fig:iterative_results_cbsd68}
\vspace{-10mm}
\end{center}
\end{figure*}

\begin{figure*}[ht]
\begin{center}
\centerline{\includegraphics[width=2.15\columnwidth]{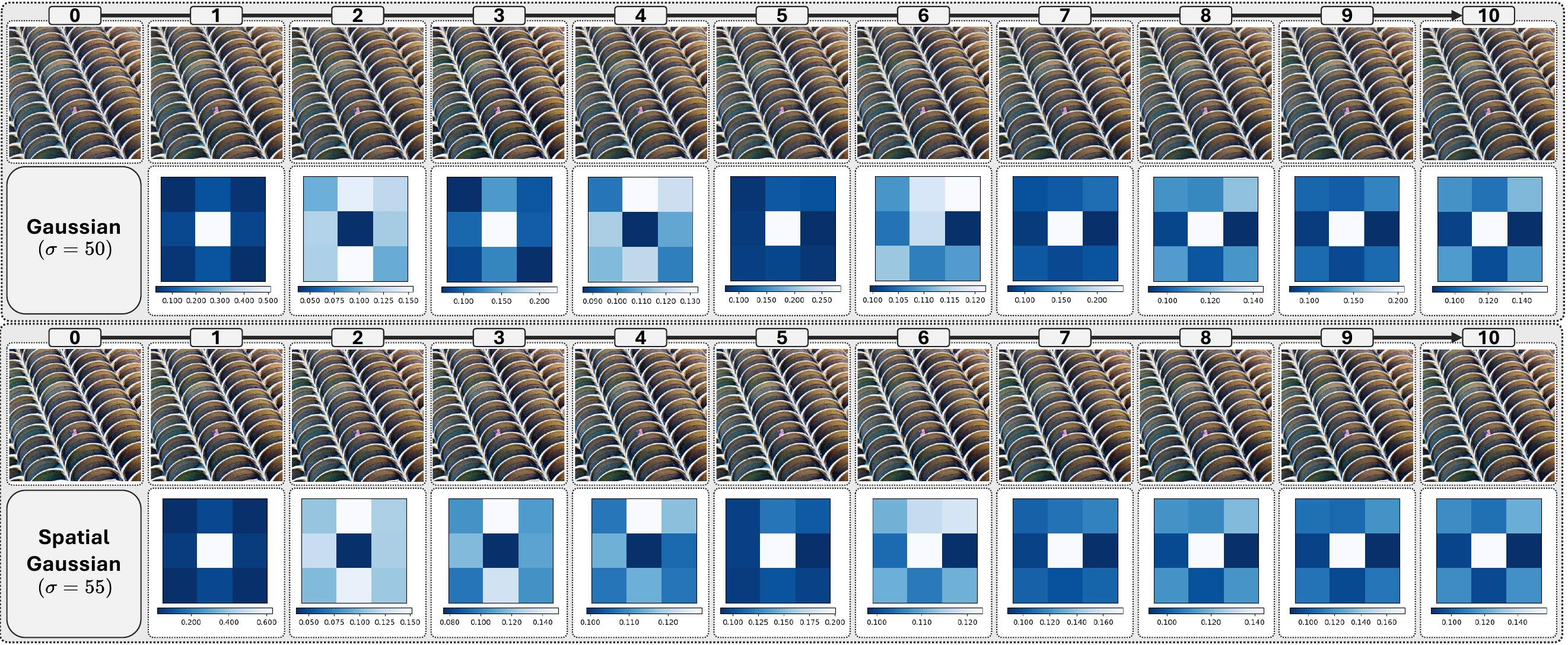}}
\caption{Comparison of results illustrating how denoising kernels and denoised images evolve through iterations, along with varying degradation types and levels on the Urban100~\cite{urban} dataset. Please zoom in for a more detailed comparison.}
\label{fig:iterative_results_urban100}
\vspace{-10mm}
\end{center}
\end{figure*}

\begin{figure*}[ht]
\begin{center}
\centerline{\includegraphics[width=2.15\columnwidth]{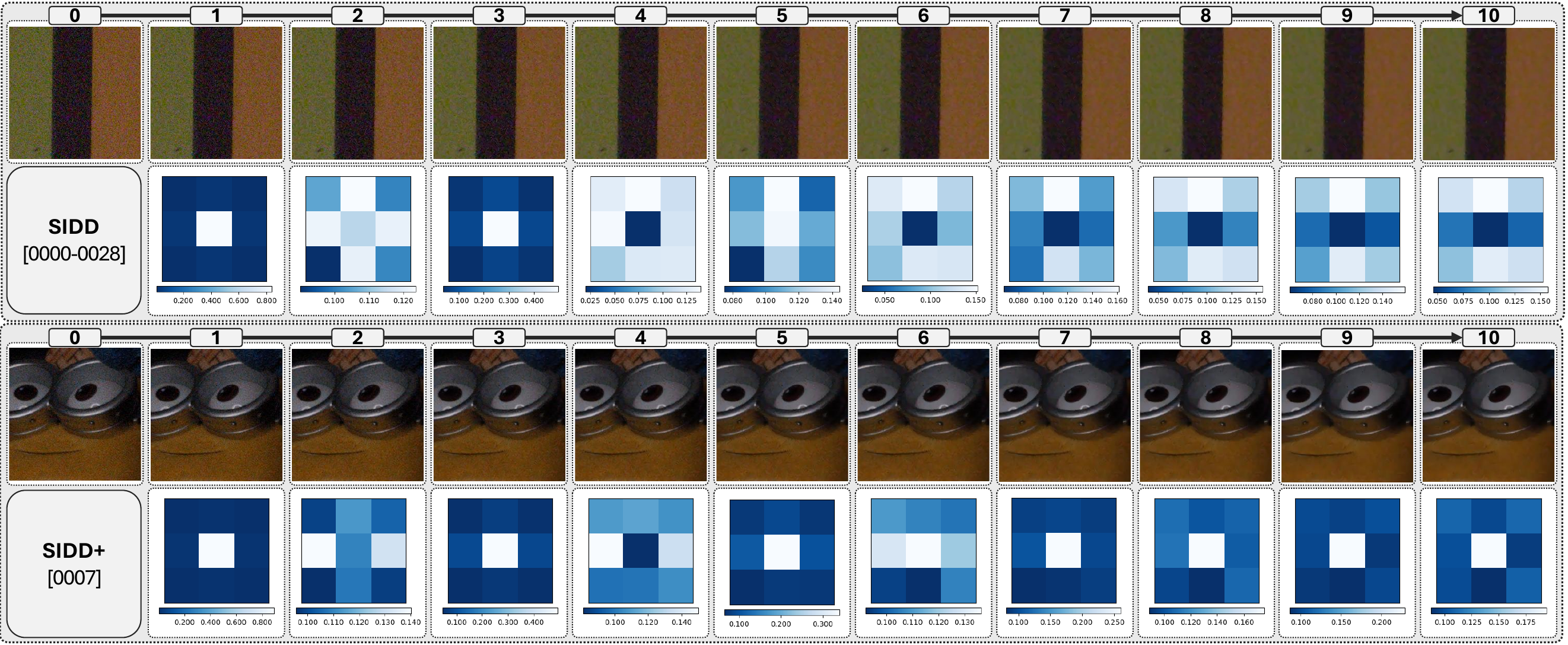}}
\caption{Comparison of results illustrating how denoising kernels and denoised images evolve through iterations, along with varying degradation types and levels on the real-world SIDD~\cite{sidd} and SIDD+~\cite{siddplus} dataset. Please zoom in for a more detailed comparison.}
\label{fig:iterative_results_real_world}
\vspace{-10mm}
\end{center}
\end{figure*}

\begin{figure*}[ht]
\begin{center}
\centerline{\includegraphics[width=2.15\columnwidth]{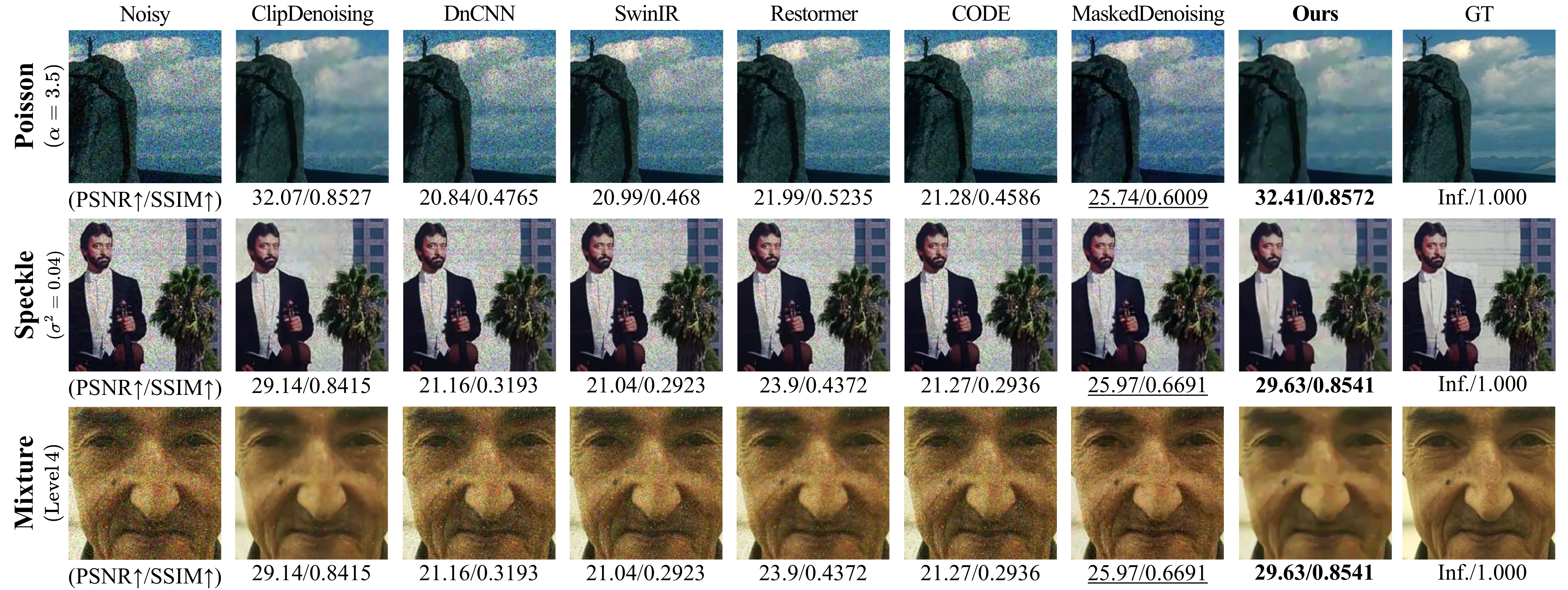}}
\caption{Qualitative results of denoising performance on synthetic OOD noise in terms of PSNR$\uparrow$/SSIM$\uparrow$. During training, none of the methods are exposed to the noise types present in the test set. Please zoom in for a more detailed comparison.}
\label{fig:abl_synthetic_denoising_extension}
\vspace{-11mm}
\end{center}
\end{figure*}

\begin{figure*}[ht]
\begin{center}
\centerline{\includegraphics[width=1.5\columnwidth]{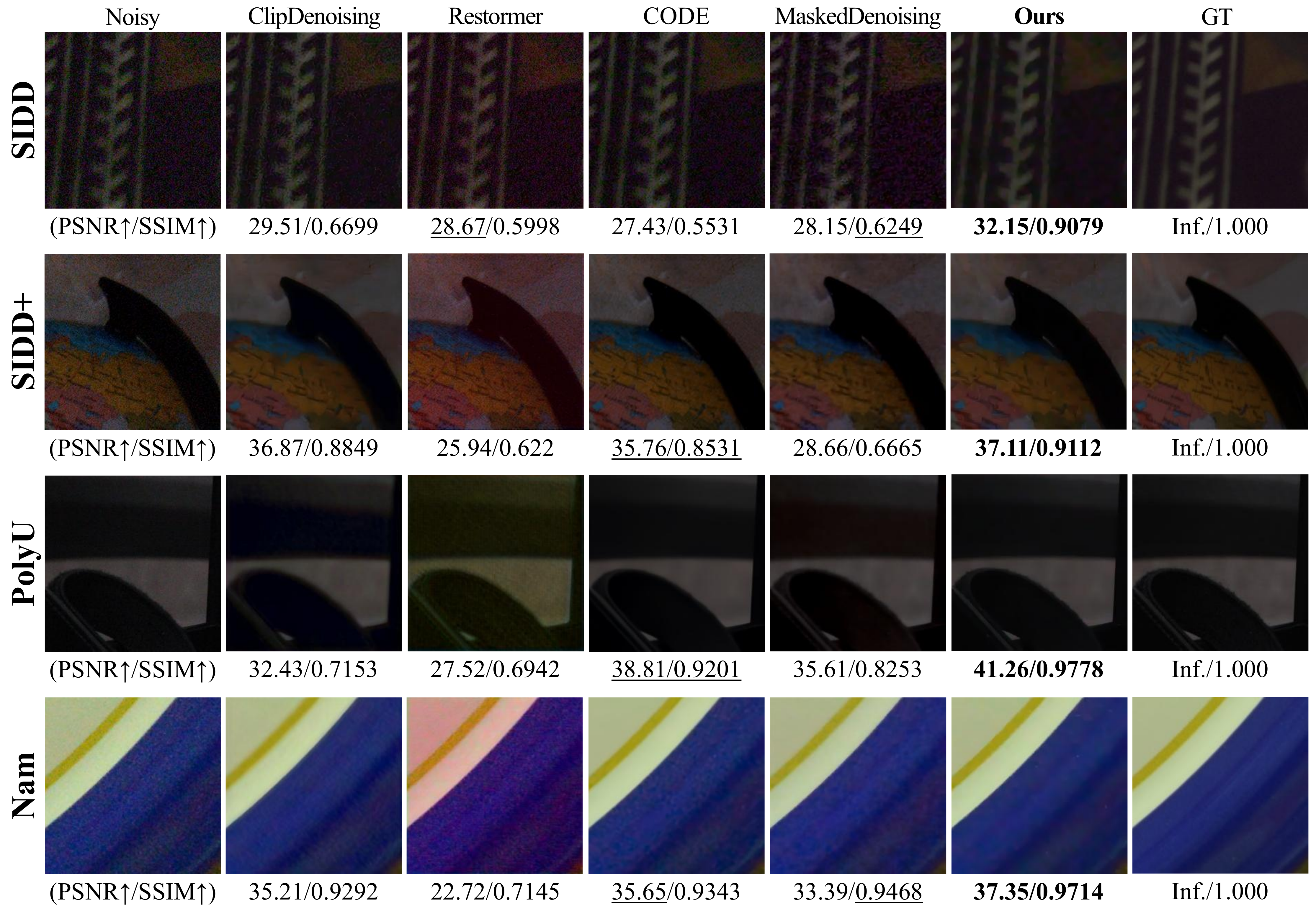}}
\caption{Qualitative results of denoising performance on real-world OOD noise in terms of PSNR$\uparrow$/SSIM$\uparrow$. During training, none of the methods are exposed to the noise types present in the test set. Please zoom in for a more detailed comparison.}
\label{fig:abl_real_world_denoising_extension}
\vspace{-11mm}
\end{center}
\end{figure*}

\begin{figure*}[ht]
\begin{center}
\centerline{\includegraphics[width=2.0\columnwidth]{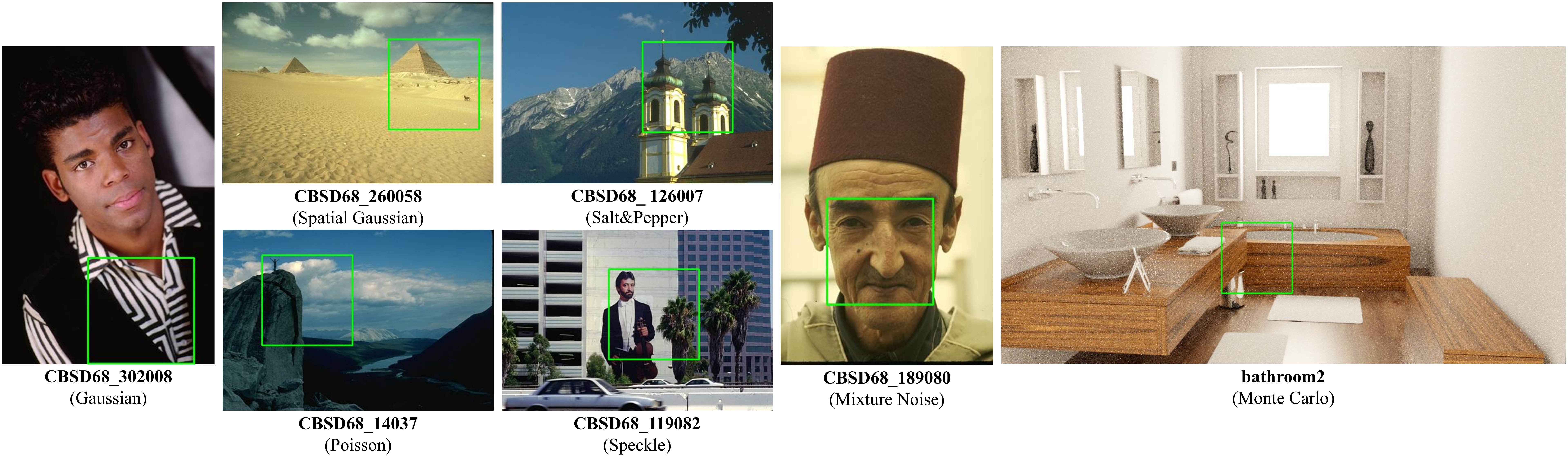}}
\caption{The original clean image and cropped region-of-interest (ROI) from the CBSD68~\cite{cbsd} and Monte Carlo rendering~\cite{mc_dataset} dataset used for the qualitative evaluation of synthetic noise removal.}
\label{fig:abl_whole_gt_image}
\vspace{-11mm}
\end{center}
\end{figure*}

\endgroup

\end{document}